\documentclass[letterpaper, 10 pt, conference]{ieeeconf}  

\IEEEoverridecommandlockouts                              

\makeatletter
\let\NAT@parse\undefined
\makeatother

\usepackage{amsmath}
\usepackage{amssymb}
\usepackage{mathtools}
\usepackage{thmtools, thm-restate}
\usepackage{wrapfig}
\usepackage{algorithm}
\usepackage{algpseudocode}
\usepackage{cite}

\usepackage{amsmath,amsfonts,bm}

\def\eqref#1{equation~\ref{#1}}

\def\1{\bm{1}}

\DeclareMathAlphabet{\mathsfit}{\encodingdefault}{\sfdefault}{m}{sl}
\SetMathAlphabet{\mathsfit}{bold}{\encodingdefault}{\sfdefault}{bx}{n}

\usepackage{graphicx}
\usepackage{booktabs} 
\usepackage{multirow}
\usepackage{multicol}
\usepackage{xspace}
\usepackage{xcolor}
\usepackage{transparent}
\usepackage{pifont}
\newcommand{\cmark}{\ding{51}}
\newcommand{\xmark}{\ding{55}}
\usepackage{caption}
\usepackage{subcaption}
\usepackage{titletoc}
\usepackage{tikz}
\usepackage{colortbl}

\usepackage{url}

\usepackage{array}
\newcolumntype{C}[1]{>{\centering\arraybackslash}m{#1}}

\usepackage{hyperref} 
\hypersetup{
    colorlinks=true,   
    linkcolor=black,   
    citecolor=blue,    
    urlcolor=blue,     
    pdfpagemode=UseNone, 
    pdfstartview=FitH, 
    bookmarksopen=true,
}
\usepackage[capitalize,noabbrev]{cleveref}

\Crefname{equation}{Eq.}{Eqs.}
\Crefname{figure}{Fig.}{Figs.}
\Crefname{table}{Tab.}{Tabs.}
\Crefname{tabular}{Tab.}{Tabs.}
\Crefname{section}{Sec.}{Secs.}
\Crefname{appendix}{App.}{Apps.}

\makeatletter
\DeclareRobustCommand\onedot{\futurelet\@let@token\@onedot}
\def\@onedot{\ifx\@let@token.\else.\null\fi\xspace}
\def\iid{\emph{i.i.d}\onedot} 
\def\eg{\emph{e.g}\onedot} 
\def\ie{\emph{i.e}\onedot}

\def\wrt{w.r.t\onedot} 
\def\aka{\emph{a.k.a}\onedot}

\makeatother

\newcommand{\modelName}{TrajFlow\xspace}

\begin{document}
\pagestyle{plain}
\title{TrajFlow: Multi-modal Motion Prediction via Flow Matching}

\renewcommand{\b}[1]{\boldsymbol{#1}}
\renewcommand{\u}[1]{\underline{#1}}


\author{Qi Yan$^{1,2}$, Brian Zhang$^{3}$, 
Yutong Zhang$^{4}$, Daniel Yang$^{5}$, Joshua White$^{6}$, Di Chen$^{7}$
\\
Jiachao Liu$^{8}$, 
Langechuan Liu$^{9}$, 
Binnan Zhuang$^{9}$, 
Shaoshuai Shi$^{10}$, 
Renjie Liao$^{1,2,11}$
\thanks{$^{1}$University of British Columbia;
        $^{2}$Vector Institute for AI;
        $^{3}$University of Waterloo;
        $^{4}$Georgia Institute of Technology; 
        $^{5}$Carnegie Mellon University;
        $^{6}$Tesla;
        $^{7}$XPeng Motors;
        $^{8}$Leapmotor;
        $^{9}$Nvidia;
        $^{10}$DiDi;
        $^{11}$Canada CIFAR AI Chair.
        Correspondence to: \texttt{rjliao@ece.ubc.ca}}
}


%

\maketitle

\begin{abstract}
Efficient and accurate motion prediction is crucial for ensuring safety and informed decision-making in autonomous driving, particularly under dynamic real-world conditions that necessitate multi-modal forecasts. 
We introduce \modelName, a novel flow matching-based motion prediction framework that addresses the scalability and efficiency challenges of existing generative trajectory prediction methods. 
Unlike conventional generative approaches that employ \iid sampling and require multiple inference passes to capture diverse outcomes, \modelName predicts multiple plausible future trajectories in a single pass, significantly reducing computational overhead while maintaining coherence across predictions. 
Moreover, we propose a ranking loss based on the Plackett-Luce distribution to improve uncertainty estimation of predicted trajectories. 
Additionally, we design a self-conditioning training technique that reuses the model’s own predictions to construct noisy inputs during a second forward pass, thereby improving generalization and accelerating inference.
Extensive experiments on the large-scale Waymo Open Motion Dataset (WOMD) demonstrate that \modelName achieves state-of-the-art performance across various key metrics, underscoring its effectiveness for safety-critical autonomous driving applications.
The code and other details are available on the project website \url{https://traj-flow.github.io/}.
\end{abstract}


\IEEEpeerreviewmaketitle

\begin{figure*}[!ht]
    \centering
    \includegraphics[width=\textwidth]{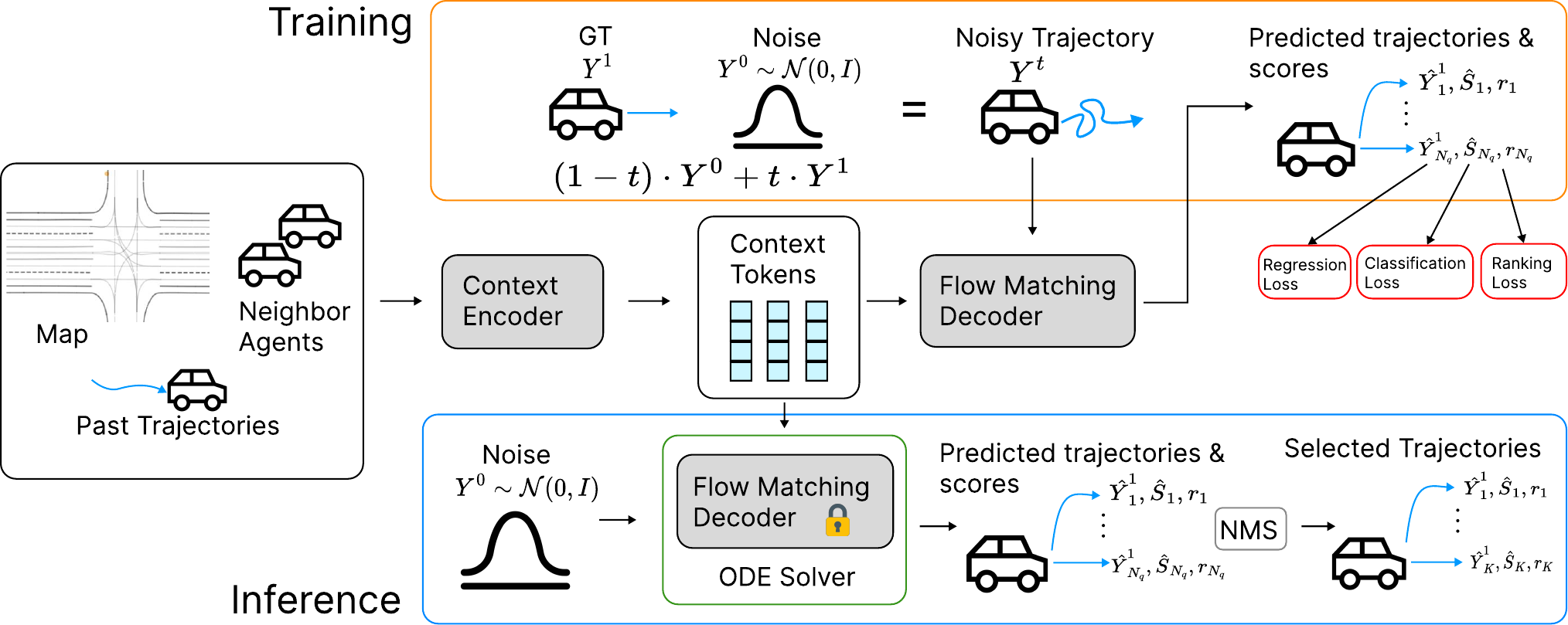}
    \caption{
    \textbf{Overview of motion prediction via flow matching.}
    The input scene, including agent history and a road map, is encoded into context tokens via a transformer. During training, noise sampled \iid from a normal distribution is added to the ground truth (GT) trajectory to create a linear interpolation. The denoiser, conditioned on these tokens, predicts denoised trajectories to align with the GT.
    During inference, trajectories are sampled from pure noise at flow time \( t = 0 \) and refined through ODEs, yielding a plausible future motion distribution.
    }
    \label{fig:overview}
    \vspace{-0.6cm}
\end{figure*}

\section{Introduction}
Motion forecasting is a fundamental component of modern autonomous driving systems and has received significant attention in recent years~\cite{liang2020learning, casas2020implicit, luo2021safety, zeng2021lanercnn, konev2022motioncnn, liu2021social, chang2019argoverse, ettinger2021large, zhao2021tnt, gu2021densetnt, Gu_2022_CVPR, Mao_2023_CVPR, jiang2023motiondiffuser, fu2025moflow,
seff2023motionlm, ngiam2021scene, nayakanti2023wayformer, varadarajan2022multipath++,
shi2022motion, shi2023mtr, zhou2023query, jia2023hdgt, lin2024eda, liu2024reasoning}. 
As a critical capability for autonomous vehicles, motion forecasting enables the prediction of future behaviors of traffic participants, including pedestrians, vehicles, and cyclists, by integrating observed agent states with road map data. 
This task is inherently challenging due to the multi-modal nature of agent behaviors and the intricate, dynamic nature of real-world traffic.

To tackle these challenges, generative models~\cite{gupta2018social, dendorfer2021mg, groupnet_CVPR, xu2024dynamic, Lee_2022_CVPR} have been widely adopted to model the distribution of future trajectories. 
Recently, diffusion models (DMs)~\cite{Gu_2022_CVPR, Mao_2023_CVPR, jiang2023motiondiffuser, fu2025moflow} have demonstrated strong capabilities in trajectory forecasting, particularly in handling spatial-temporal dependencies. 
However, despite their success, existing generative motion prediction models face critical challenges that hinder their deployment in large-scale, real-world applications.
First, conventional generative models rely on an \iid sampling process, producing one trajectory per run and requiring multiple sampling passes to capture multi-modal behaviors, thereby increasing computational overhead. 
Furthermore, independently generated trajectories may exhibit inconsistencies, undermining the coherence and reliability of multi-modal predictions.
Second, diffusion models suffer from slow inference due to their iterative denoising process, making them impractical for real-time autonomous driving applications, where rapid decision-making is crucial.

To address these limitations, we introduce \textit{\modelName}, a flow-matching-based model designed to efficiently learn the distribution of future multi-modal trajectories. 
Our approach builds on flow matching, a broader class of generative models encompassing DMs~\cite{ lipman2022flow}. 
Conditioned on the historical context, \modelName learns to map noise vectors to plausible future trajectories by solving ordinary differential equations (ODEs).
We summarize our key contributions as follows:  
\begin{itemize}
    \item We introduce a novel multi-shot flow matching framework that jointly predicts multiple plausible future trajectories in a single sampling pass, eliminating the inefficiencies of \iid sampling. This approach significantly reduces computational cost while ensuring coherent and diverse multi-modal predictions.
    \item We propose a Plackett-Luce distribution-based ranking loss to better calibrate the uncertainty of predicted trajectories, improving the reliability for downstream tasks such as trajectory selection and decision-making.
    \item We present a self-conditioning training strategy that mitigates overfitting in flow matching when noisy inputs closely resemble targets. This technique enhances model generalization while accelerating inference, enabling more efficient real-time predictions.
    \item Extensive experiments on the large-scale Waymo Open Motion Dataset (WOMD) demonstrate that \modelName achieves state-of-the-art performance across various key metrics, verifying its effectiveness.
\end{itemize}

\section{Related Work}

\noindent\textbf{Motion Forecasting Models.}
Large-scale datasets like the Waymo Open Motion Dataset (WOMD)~\cite{ettinger2021large} have significantly accelerated the development of motion prediction models. These models effectively fuse multi-modal inputs, capturing static, dynamic, social, and temporal elements of the scene, essential for understanding complex scene data and traffic agent interactions. 
Drawing inspiration from computer vision, one class of models visualizes inputs as multichannel rasterized top-down images~\cite{casas2020implicit, luo2021safety, zeng2021lanercnn, konev2022motioncnn}, leveraging spatio-temporal convolutional networks to represent relationships in an orthographic plane. More recently, transformer-based models with encoder-decoder architectures have become prevalent~\cite{seff2023motionlm, ngiam2021scene, jiang2023motiondiffuser, nayakanti2023wayformer, varadarajan2022multipath++,
shi2022motion, shi2023mtr, zhou2023query, jia2023hdgt, lin2024eda, liu2024reasoning}, focusing on agent-to-environment interaction through attention layers~\cite{vaswani2017attention}.
The multi-modal nature of trajectory prediction is essential for accurate motion forecasting. Recent models typically generate a set of candidate trajectories, which are then pruned for deployment. They often use Non-Maximum Suppression (NMS)~\cite{shi2022motion, shi2023mtr, zhou2023query, jia2023hdgt, lin2024eda, liu2024reasoning} to filter trajectories based on endpoints and confidence scores.

\noindent\textbf{Denoising Models for Motion Prediction.}  
Driven by the success of diffusion~\cite{ho2020denoising} and flow matching~\cite{lipman2022flow}, researchers have recently focused on developing diffusion models for motion prediction~\cite{Gu_2022_CVPR, Mao_2023_CVPR, jiang2023motiondiffuser, fu2025moflow}.  
A common drawback of these denoising models is their slow sampling process, necessitating student model distillation~\cite{Mao_2023_CVPR, fu2025moflow} or advanced sampling techniques~\cite{jiang2023motiondiffuser}.
Naively building conditional generative models with an \iid sampling strategy incurs high computational costs~\cite{jiang2023motiondiffuser}. Moreover, independent sampling struggles to produce coherent trajectories and confidence scores for multi-modal predictions. Diffusion models further exacerbate this by making likelihood evaluation expensive, complicating both waypoint generation and confidence estimation steps.
Interestingly, diffusion models have also been applied to other motion-related tasks, such as human motion~\cite{chen2023executing, tevet2022human, zhang2024motiondiffuse} and hand-object interaction~\cite{liu2024geneoh, lee2024interhandgen, Li_2025_CVPR}. While these areas fall outside the scope of this paper, they underscore the expressive power of diffusion modeling.

\section{Method}
In this section, we introduce the proposed \modelName model. 
We first cover preliminaries on motion prediction and flow matching. 
Then, we present our model design, including a novel flow matching framework for multi-modal trajectory prediction. 
The overall structure is illustrated in~\cref{fig:overview}.

\subsection{Preliminaries}
\label{sect:method_prelim}
\par\noindent\textbf{Motion Prediction.}
Given a driving scene, we consider the agent-centric setup for motion prediction.
Let the state of the ego agent (\aka, agent of interest) be \(X \in \mathbb{R}^{T_p \times D_a}\), where \(T_p\) is the number of frames in the history, and \(D_a\) is the state dimension (\eg, positions and velocity). 
We denote the set of neighboring agents \wrt the ego agent as \(\mathcal{N}\) and their states as \(X_{\mathcal{N}} \in \mathbb{R}^{\vert \mathcal{N} \vert \times T_p \times D_a}\).
Let $N_a = \vert \mathcal{N} \vert + 1$ be the total number of agents.
Let \(M \in \mathbb{R}^{N_m \times D_p \times D_m}\) denote the set of road map polylines in the given scene, where \(N_m\) is the number of polylines, \(D_p\) is the number of points per polyline, and \(D_m\) is the number of attributes (\eg, location and type). 
For brevity, we let \(C := \{X, X_{\mathcal{N}}, M\}\) denote the contextual information of the ego agent.
Our aim is to predict $K$ multi-modal future trajectories \(Y_1, Y_2, \cdots, Y_K\) with corresponding confidence scores $\eta_1, \eta_2, \cdots, \eta_K$ for the ego agent. 
Each prediction $Y_i$ is in the shape of  \(\mathbb{R}^{T_f \times D_t}\), where \(T_f\) is the number of future timesteps, and \(D_t\) is the dimension (often chosen as 2 to represent X-Y coordinates).

\par\noindent\textbf{Flow Matching.}
We adopt the simple linear flow matching model from~\cite{lipman2022flow}, which learns ordinary differential equations (ODEs) to transform noise into data.
Following the flow matching notation, let $Y^0 \sim \mathcal{N}(\mathbf{0}, \mathbf{I})$ and $Y^1$ be noise and observed future trajectories, respectively.
To construct the noisy input $Y^t$, we linearly interpolate between $Y^1$ and $Y^0$, where $t \in [0, 1)$ represents the flow model time, distinct from the trajectory timestamp.
Specifically, we define:
\begin{align}
    Y^t &= (1-t)\,Y^0 + t\,Y^1, 
    \label{eq:flow_interpolation_1}
    \\
    V^t &= Y^1 - Y^0,
    \label{eq:flow_interpolation_2}
\end{align}
with the velocity field \( V^t \) guiding the probability flow from noise to data distribution.
Although \(V^t\) depends on \(t\) in general, linear flow assumes a constant velocity for simplicity. 
During training, we sample \( t \) from a predefined distribution, \eg, uniform, and train a neural network \(v_\theta\) to approximate \(V^t\) by minimizing
\begin{align}
    L_{\text{flow}} := \mathbb{E}_{Y^t, Y^1, t} \left[ \Vert v_\theta(Y^t, C, t) - V^t \Vert^2_2 \right].
    \label{eq:loss_fm_raw}
\end{align}
Here, \(v_\theta\) also incorporates the aforementioned context information \(C\).
During inference, we begin with \(Y^0 \sim \mathcal{N}(\mathbf{0}, \mathbf{I})\) and iteratively update it toward \(t = 1\) via
\(
Y^{t+1} = Y^t + v_\theta(Y^t, C, t),
\)
which essentially solves the underlying ODE using the Euler’s method.

\subsection{Flow Matching for Multi-modal Motion Prediction}
\label{sec:method_fm}
In this study, we propose a novel framework to model the multi-modal future trajectories using flow matching models. 

\subsubsection{Context Encoder}
In this paper, we utilize agent-centric representations~\cite{varadarajan2022multipath++, shi2023mtr}, where all inputs are transformed into a coordinate system centered on the agent.
We employ a PointNet~\cite{qi2017pointnet} architecture to encode agent and map elements, generating tokenized features \(Z_a = \phi(\mathrm{MLP}(X)) \in \mathbb{R}^{N_a \times D}\) and \(Z_m = \phi(\mathrm{MLP}(M)) \in \mathbb{R}^{N_m \times D}\), where \(\mathrm{MLP}\) is a linear network and \(\phi\) denotes a max-pooling operator.  
Following MTR~\cite{shi2022motion}, we apply multiple multi-head attention (MHA) layers~\cite{vaswani2017attention} to iteratively refine the context tokens, initialized as \(Z_{s}^0 = \texttt{cat}(Z_a, Z_m) \in \mathbb{R}^{(N_a + N_m) \times D}\) after concatenation.  
At the \(j\)-th attention layer, the context tokens are updated as  
\begin{align*}
    \resizebox{0.95\columnwidth}{!}{$
    Z_{s}^{j} \!=\! \mathrm{MHA}(\overbrace{Z_{s}^{j-1} \!+\! \mathrm{PE}_{Z_{s}^{j-1}}}^{\mathrm{Query}}, \overbrace{\mathrm{Local} (Z_{s}^{j-1} \!+\! \mathrm{PE}_{Z_{s}^{j-1}})}^{\mathrm{Key}},  \overbrace{Z_{s}^{j-1}}^{\mathrm{Value}}),
    $}
\end{align*}
where \(\mathrm{PE}\) is the location-based sinusoidal positional encoding, and \(\mathrm{Local}(\cdot)\) restricts attention to the nearest neighbors based on their last history frame's Euclidean distance.
Finally, the encoder outputs the refined context tokens \(Z_s\), which serve as inputs for the cross-attention layers in the decoder.
Note that the above representations are agent-centric and apply to a single agent at a time. As a result, there are \(N_a\) copies of the context tokens \(Z_s\), one for each agent. 

\subsubsection{Trajectory-Space Flow Matching Decoder}\label{subsubsect:decoder}
\begin{figure}[t]
    \centering
    \includegraphics[width=0.8\linewidth]{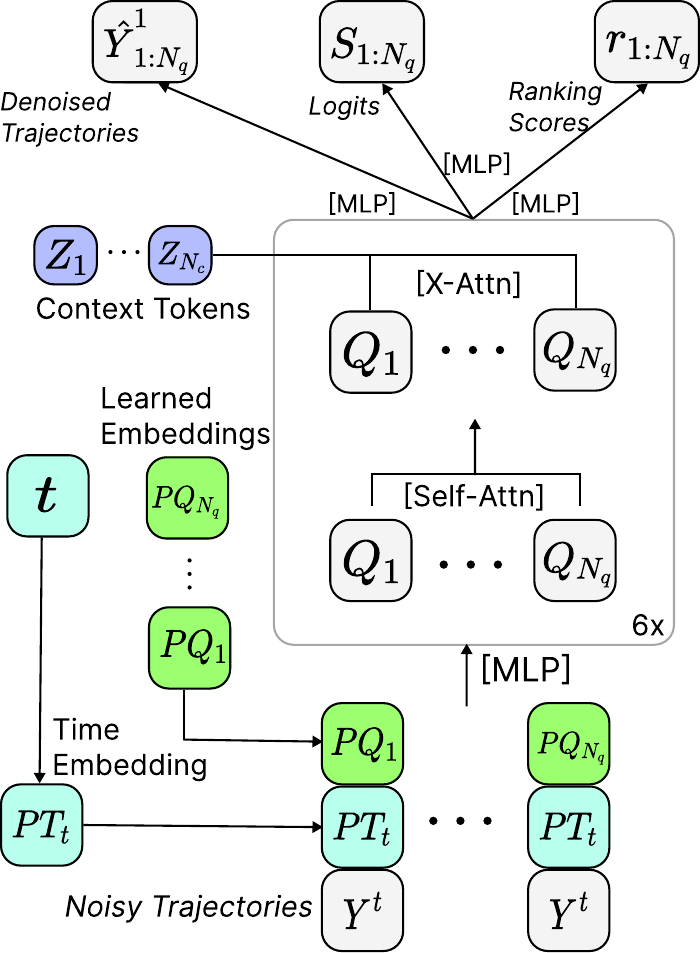}
    \caption{
    Network architecture of flow matching decoder.
    }
    \label{fig:architecture}
    \vspace{-0.5cm}
\end{figure}

We adopt the query-based attention decoder design~\cite{jiang2023motiondiffuser, nayakanti2023wayformer, varadarajan2022multipath++,
shi2022motion, shi2023mtr, zhou2023query, jia2023hdgt, lin2024eda, liu2024reasoning}, as shown in Fig.~\ref{fig:architecture}.  
To enable multi-modal trajectory prediction, we generate \(N_q\) query tokens per agent.  
The decoder then takes as input the flow time \(t\) and the noise-corrupted trajectory \(Y^t\).
For the $i$-th query, its initial token is \(Q_i^0 = \mathrm{MLP}\left(\texttt{cat}(\mathrm{MLP}(Y^t), \mathrm{PT}_t, \mathrm{PQ}_i)\right) \in \mathbb{R}^D\),
where \(\mathrm{PT}_t\) is the flow time positional embedding, and \(\mathrm{PQ}_i\) is a learnable positional embedding that projects the context token of the center agent, which is one element of the agent tokens \(Z_a\).  
The full set of initial queries is given by  
$Q^0 = \texttt{cat} (Q_1^0, \dots, Q_{N_q}^0) \in \mathbb{R}^{N_q \times D}$.
Next, the query tokens go through multiple layers of inter-query self-attention and query-to-context cross-attention, defined at the \(j\)-th layer as  
\begin{align*}
    Q^{j}_{sa} &\!=\! \mathrm{MHA}({Q^{j-1} + \mathrm{PQ}}, {Q^{j-1} + \mathrm{PQ}},  {Q^{j-1}}),
    \\
    Q^{j} &\!=\! \mathrm{MHA}( {\texttt{cat} (Q^{j}_{sa}, \mathrm{PQ})}, {\texttt{cat}(Z_s, \mathrm{PE}_{Z_s})},  {Z_s}).
\end{align*}
We incorporate \(\mathrm{PQ}\) embeddings for self-attention positional encoding in the query and key projections, as well as in the cross-attention query projections.  
To enhance spatial reasoning, we introduce location-based positional encoding \(\mathrm{PE}_{Z_s}\) in the cross-attention layer.  
For efficiency, we employ a local attention mechanism in the cross-attention, restricting interactions to the nearest context token relative to the central agent's last observable position.  
The final output is projected by three MLP heads to produce $N_q$ denoised trajectories \(\hat{Y}^1 \in \mathbb{R}^{N_q \times T_f \times D_t}\) and their corresponding logits (unnormalized scores) \(S \in \mathbb{R}^{N_q}\).  
Denoting the whole encoder-decoder model as $F_\theta$, we have \(\hat{Y}^1, S = F_\theta(Y^t, C, t)\).
Our joint trajectory prediction approach differs from standard conditional generation~\cite{jiang2023motiondiffuser}, which produces only one trajectory at a time. Our novel training objective, detailed below, enables the generation of multiple trajectories in a single sampling run, greatly reducing computation cost during inference.

\subsubsection{Loss Functions}
We now introduce the loss functions of our model, \ie, the multi-modal trajectory prediction losses and the learning-to-rank loss. 

\par\noindent\textbf{Multi-modal Trajectory Prediction Losses.}
To encourage the prediction of multi-modal future trajectories, we replace the velocity-space flow matching loss in Eq.~\ref{eq:loss_fm_raw} with a trajectory-space loss. 
Recall the model $F_{\theta}$ predicts $N_q$ future trajectories $Y^1$ and their corresponding logits $S$, where the $k$-th trajectory and its logit are given by \(Y^1_k \in \mathbb{R}^{T_f \times D_t}\) and \(S_k\), respectively.
We first identify the best trajectory as the one closest to the ground truth trajectory \( Y^1 \),
\[k^{\ast} = \textstyle \arg \min_k \Vert \hat{Y}^1_{k} - Y^1 \Vert_2^2.\]

Before computing the \( \arg\min \) operator, we apply Non-Maximum Suppression (NMS) based on trajectory logits, following~\cite{lin2024eda}. 
NMS leverages trajectory logits and an endpoint distance threshold to suppress redundant trajectories, generating a fixed number of trajectories from the $N_q$ proposals.
Its goal is to select trajectories with representative and diverse motion patterns.
We then employ a regression loss to encourage the best trajectory to closely match the ground truth and a classification loss to maximize its score while minimizing the scores of others.
Specifically, we define,
\begin{align}
    L_{\text{flow}} &= \mathbb{E}_{Y^t, Y^1, t} \left[ \textstyle  \Vert \hat{Y}^1_{k^{\ast}} - Y^1 \Vert_2^2 \right], \label{eq:loss_fm_reg} \\
    \textstyle L_{\text{cls}} &= \mathbb{E}_{Y^t, Y^1, t} \left[\textstyle 
 \frac{1}{N_q} \sum_{k=1}^{N_q} \mathrm{BCE}(S_{k}, \mathbf{1}_{k=k^{\ast}}) \right],   
 \label{eq:loss_fm_cls}
\end{align}
where \( \mathrm{BCE} \) denotes the binary cross-entropy loss, and \( \mathbf{1} \) is the indicator function.
The confidence score of the \(k\)-th predicted trajectory is given by \(\hat{S}_k = \mathrm{sigmoid}(S_k)\).  
To improve training stability, we adopt the Gaussian Mixture Model (GMM) formulation for the regression loss, following \cite{shi2022motion}.  
Further details can be found in the original reference.

\par\noindent\textbf{Learning-to-Rank Loss.}
The confidence scores $\hat{S}$ of predicted trajectories are not necessarily calibrated, \ie, a higher score does not always correspond to better accuracy.  
To promote calibration, we introduce a ranking loss based on the Plackett-Luce (PL) distribution~\cite{plackett,luce, zhang2024symmetricdiffusers}.  
To this end, we add a prediction head in \(F_\theta\) to assign ranking scores \(\{r_{k} \mid k = 1, \dots, N_q \}\), where \(r_{k}\) represents the preference for $k$-th prediction for the ego agent.  
We define the ground-truth ranking as \(\sigma = \arg \mathrm{sort}( [d_1, \dots, d_{N_q}] )\), where \(d_k = \Vert \hat{Y}^1_{k} - Y^1 \Vert_2^2\) is the displacement error.  
Under the PL model parameterized by $\{r_k\}$, the probability of the ground-truth ranking \(\sigma\) is given by:
\begin{align}
    p_{\mathrm{PL}}(\sigma) = \prod_{k=1}^{N_q} \frac{\exp(r_{\sigma(k)})}{\sum_{j=k}^{N_q} \exp (r_{\sigma(j)})}.
\end{align}
The PL distribution implies the following sequential sampling process:  
1) Sample the first rank \(\pi(1)\) from \(\mathrm{Cat}(N_q, \mathrm{softmax}(r_{1:N_q}))\).  
2) Remove \(\pi(1)\) and sample \(\pi(2)\) from a categorical distribution constructed using remaining ranking scores.  
3) Continue until all ranks \(\pi(1), \ldots, \pi(K)\) are assigned.  
Here $\mathrm{Cat}(N, p)$ denotes the categorical distribution with $N$ outcomes and probabilities $p$.
Therefore, it is natural to define the ranking loss as the negative log likelihood of the PL model,  
\begin{align}
    L_{\text{rank}} = - \log p_{\mathrm{PL}}(\sigma). \label{eq:loss_rank}
\end{align}
In practice, we use another set of scores, \( r_{1:N_q} \), instead of the BCE scores \( \hat{S}_{1:N_q} \) to compute the PL loss, ensuring training stability. Empirically, we find that using BCE scores in the PL loss leads to overly strong regularization and results in worse performance.
The total training loss, incorporating the previously discussed flow matching regression loss and confidence score classification loss, is given by  
\begin{align}
    L = L_{\text{flow}} + L_{\text{cls}} + \lambda  L_{\text{rank}}.
    \label{eq:loss_fm_total}
\end{align}
We set $\lambda = 0.1$ in our implementation to boost performance.

\subsubsection{Self-Conditioning to Prevent Overfitting}\label{sect:method_overfit}
We observe that the introduced trajectory-space decoder in Section \ref{subsubsect:decoder} may overfit when \(t\) is close to 1, as \(Y^t\) remains highly similar to the ground truth \(Y^1\) due to the linear interpolation in~\cref{eq:flow_interpolation_1}.  
This allows the model \(F_\theta\) to exploit a trivial solution by simply outputting values close to its input, without effectively leveraging the context \(C\), which significantly degrades inference performance.  
To mitigate this, we introduce a self-conditioning mechanism that leverages the model's own predictions to construct noisy inputs.  
With a 50\% probability, the model first predicts trajectories at \(t=1\), which are then used in~\cref{eq:flow_interpolation_1} to generate noisy inputs at a different stochastic time \(t\). 
These inputs are then fed back into the model for a second round of the denoising process.  
This prevents the model from relying solely on ground-truth trajectories for noisy sample generation, reducing the risk of overfitting, as validated in~\cref{sec:exp_ablation}.

\subsubsection{Inference}
As described in~\cref{sec:method_fm}, each agent generates \(N_q\) trajectory outputs to capture the multi-modal nature of dynamic motion.  
Prior works~\cite{shi2022motion, shi2023mtr, lin2024eda, liu2024reasoning} suggest using a larger \(N_q\) to better cover diverse motion patterns and reflect real-world dynamics. 
However, \(N_q\) often exceeds \(K\), the required number of trajectories for evaluation.  
To address this, we apply the commonly used Non-Maximum Suppression (NMS) approach to select \(K\) trajectories per agent.  
NMS leverages confidence scores \(\mathrm{sigmoid}(S_k)\) and an endpoint distance threshold to suppress redundant trajectories, ensuring that the final selection remains both representative and diverse.
We use the hyper-parameters from~\cite{lin2024eda} to implement NMS, following the same pipeline as during training.

\begin{figure*}
    \centering
    \includegraphics[width=0.48\textwidth]{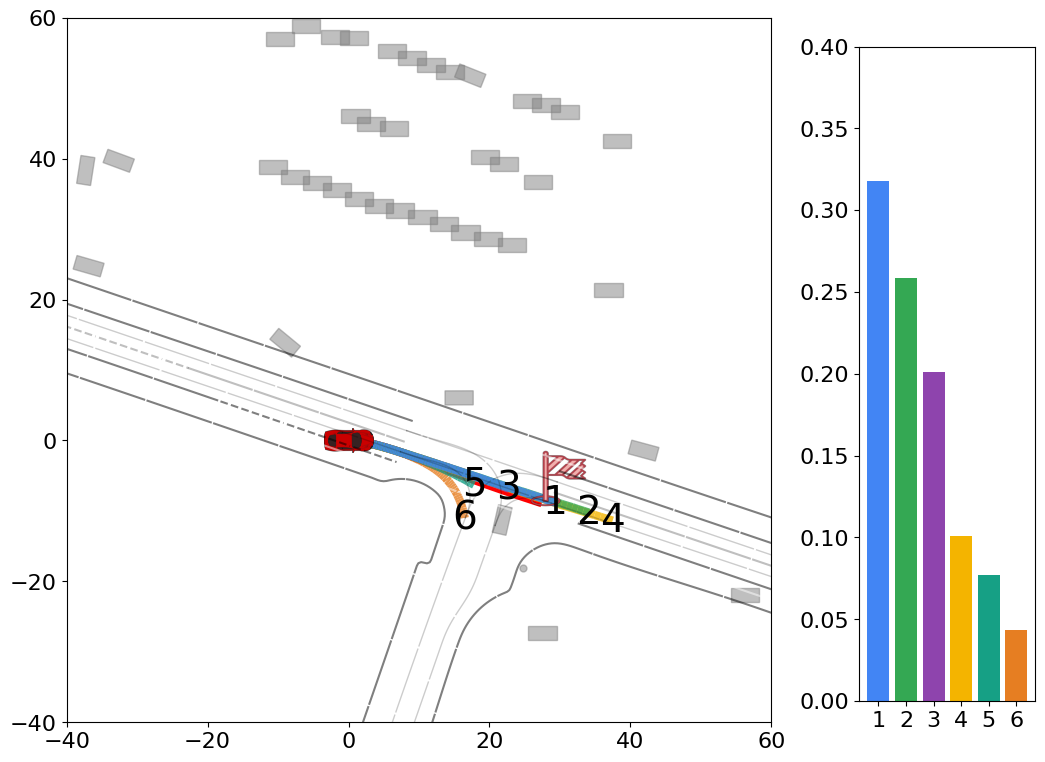} 
    \includegraphics[width=0.48\textwidth]{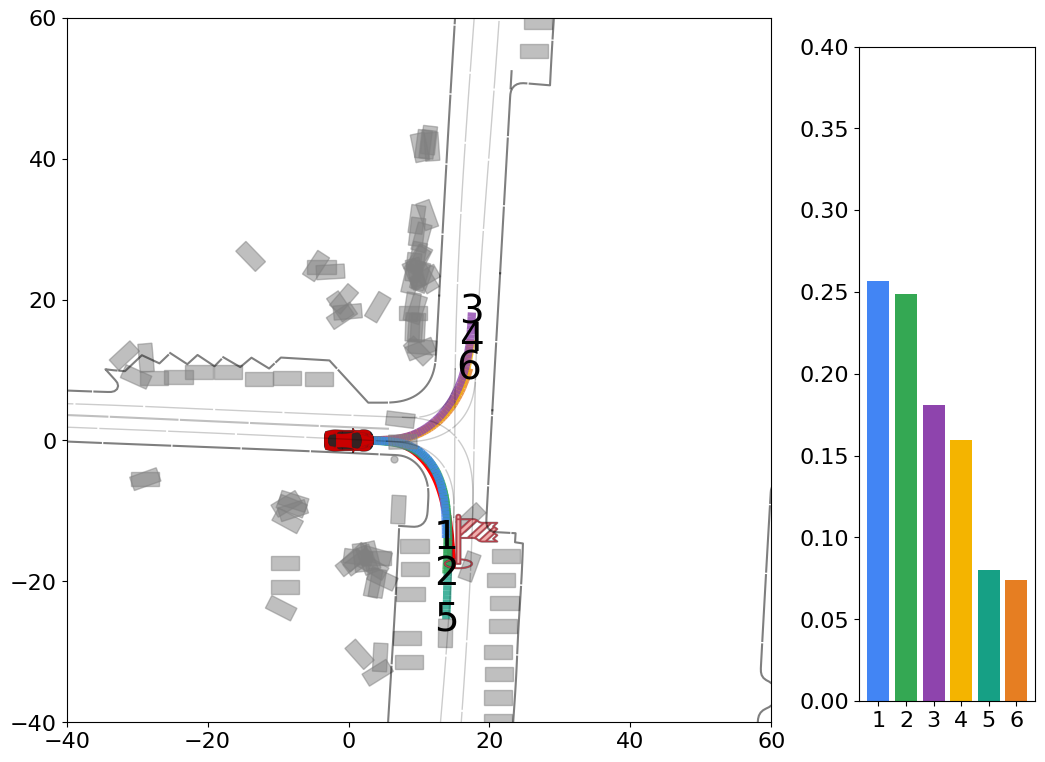}
    \includegraphics[width=0.48\textwidth]{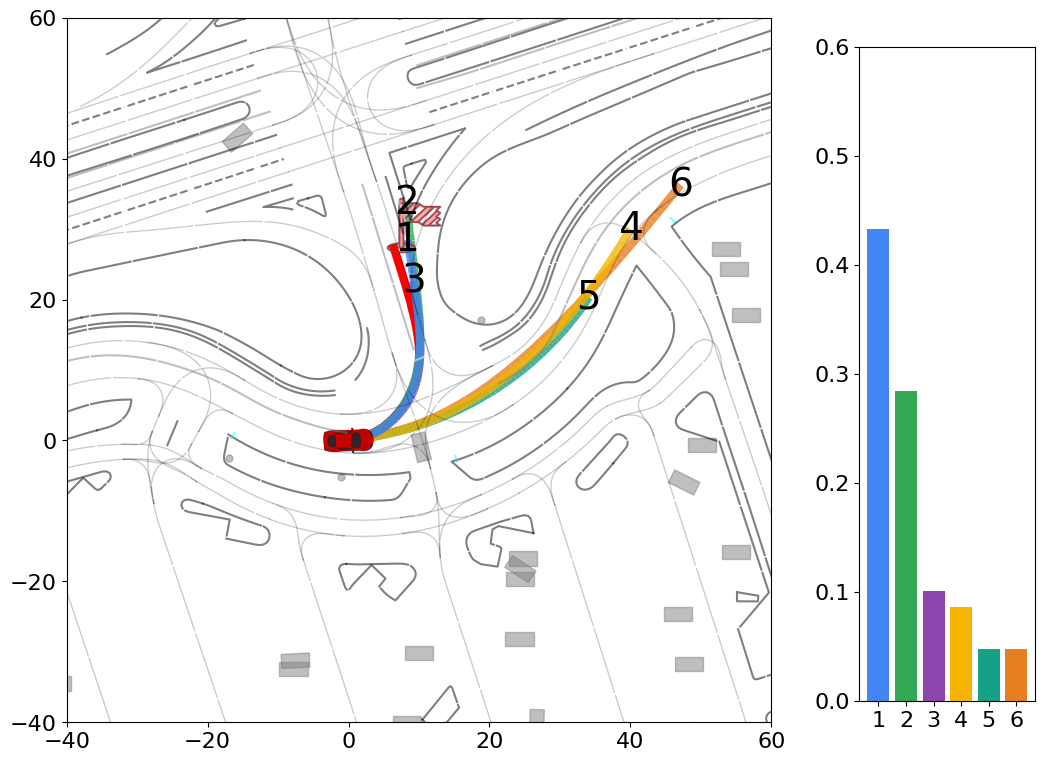}
    \includegraphics[width=0.48\textwidth]{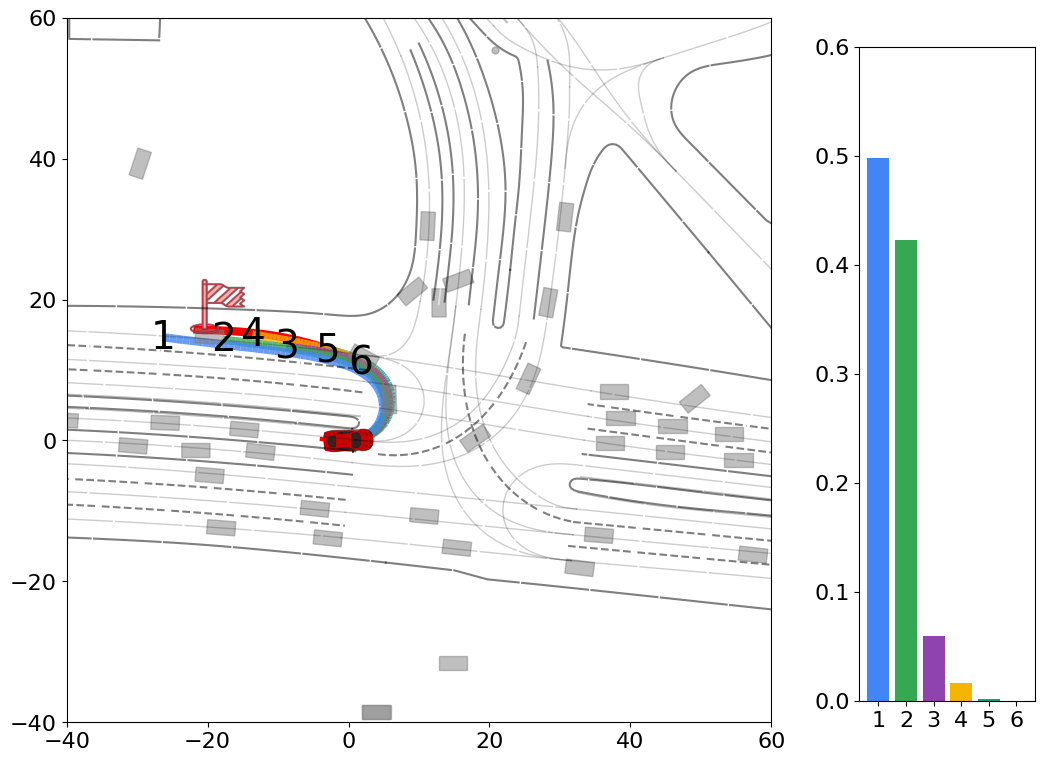}
    \caption{
    \textbf{Qualitative results.} 
    We visualize trajectory predictions (left) and their normalized confidence scores (right) in each plot. Six trajectory predictions are marked with different colors and numbered at their endpoints; the most confident prediction is shown in blue. The ground truth is displayed in red, with a flag indicating its endpoint. Surrounding road elements, such as lanes and sidewalks, are shown as lines and dots. Other surrounding traffic objects are represented as gray boxes.
    }
    \label{fig:traj_success}
    \vspace{-0.6cm}
\end{figure*}


\begin{table}[t]
\caption{\textbf{Performance of marginal prediction on WOMD testing set. }
For a fair comparison, we compare our method with existing motion prediction approaches without using model ensembling or extra data such as LIDAR.
}
\vspace{-0.1cm}
\centering
\label{tab:results_womd_motion}
\begin{tabular}{l@{\hspace{2pt}}|c@{\hspace{6pt}}c@{\hspace{6pt}}c@{\hspace{6pt}}|c@{\hspace{6pt}}c@{\hspace{4pt}}}
\toprule
Method & minADE$\downarrow$ & minFDE$\downarrow$ & \begin{tabular}[l]{@{}>{}l@{}}Miss \\Rate\end{tabular}$\raisebox{0.0ex}{$\downarrow$}$ & mAP$\uparrow$ & \begin{tabular}[c]{@{}>{}l@{}}Soft \\mAP\end{tabular}$\raisebox{0.0ex}{$\uparrow$}$ \\ 
\midrule
MotionCNN~\cite{konev2022motioncnn} & 0.7400 & 1.4936 & 0.2091 & 0.2136 & - \\
ReCoAt~\cite{huang2022recoat} & 0.7703 & 1.6668  & 0.2437 & 0.2711 & - \\
DenseTNT~\cite{gu2021densetnt} & 1.0387 & 1.5514 & 0.1573 & 0.3281 & - \\
SceneTF~\cite{ngiam2021scene} & 0.6117 & 1.2116 & 0.1564 & 0.2788 & - \\
HDGT~\cite{jia2023hdgt} & 0.5933 & 1.2055 & 0.1854 & 0.3577 & 0.3709 \\
MTR~\cite{shi2022motion} & 0.6050 & 1.2207 & 0.1351 & 0.4129 & 0.4216 \\ 
MTR++~\cite{shi2023mtr} & 0.5906 & 1.1939 & 0.1298  & 0.4329 & 0.4414 \\  
EDA~\cite{lin2024eda} & \underline{0.5718} & {1.1702}  & \underline{0.1169}  & {0.4487} & {0.4596} \\ 
CtrlMTR~\cite{sun2024controlmtr} & 0.5897 & 1.1916 & 0.1282  & 0.4414 & 0.4572 \\ 
{BeTop}~\cite{liu2024reasoning} & {0.5723} & \underline{1.1668} & {0.1176} & \underline{0.4566} & \underline{0.4678} \\ 
\midrule
\textbf{\modelName} & \textbf{0.5714} & \textbf{1.1667} & \textbf{0.1162} & \textbf{0.4604} & \textbf{0.4710} \\ 
\bottomrule
\end{tabular}
\vspace{-0.5cm}
\end{table}

\section{Experiments}
\subsection{Experimental Setup}
\subsubsection{Datasets}
We evaluate our method on the \textbf{Waymo Open Motion Dataset} (WOMD)~\cite{ettinger2021large}, a widely recognized benchmark covering diverse real-world traffic scenarios.  
Models are given 1 second of historical data to predict \(K=6\) distinct trajectories, each extending 8 seconds into the future.  
The dataset consists of approximately 487K training instances, 44K validation instances, and 44K testing instances.  
We demonstrate competitive performance on both marginal and joint motion prediction tasks.

\subsubsection{Metrics}
We evaluate performance using multiple metrics, including minimum Average Displacement Error (\textbf{minADE}), minimum Final Displacement Error (\textbf{minFDE}), and \textbf{Miss Rate}.  
minADE computes the average \(\ell_2\) distance between the ground-truth trajectory and the closest of \(K=6\) predictions across all timesteps, while minFDE measures the final timestep error, emphasizing long-term accuracy.  
Miss Rate quantifies the proportion of cases where minFDE exceeds a threshold, indicating significant deviation.  
Additionally, we use mean Average Precision (\textbf{mAP}) to assess multi-modal prediction precision.  
To compute mAP, ground truth trajectories are grouped by motion type into buckets. For each bucket, predictions are ranked by confidence and labeled true or false positives, counting only the highest-confidence true positive per object while treating others as false positives. Precision-recall curves are computed using interpolated precision, averaging across buckets for the final mAP.
\textbf{Soft mAP} is similar but does not penalize multiple matching predictions beyond the highest-confidence one.
Both mAP metrics require prediction scores to preserve relative ranking for accurate evaluation.
For further details, please refer to the official leaderboard~\cite{ettinger2021large}.

\subsubsection{Implementation Details}
We design an encoder with \(6\) transformer layers, each with a token dimension of \(256\), to model traffic scenes.  
Map information is represented by \(N_m = 768\) polylines, utilizing a local attention window of \(16\), while data from all neighboring agents are incorporated to encode the traffic context.  
Our decoder has \(6\) transformer layers, each with a token dimension of \(512\).  
We set \(N_q = 64\) to generate a diverse pool of plausible trajectories for NMS selection.  
The query-to-context local attention mechanism limits the context length to \(384\).  
Training is conducted on \(8\) NVIDIA A100 (80GB) GPUs using the AdamW optimizer~\cite{loshchilov2017decoupled} within the PyTorch framework~\cite{paszke2019pytorch}.  
The model is trained for \(40\) epochs with a learning rate of \(0.0001\) (linearly decayed) and a weight decay of \(0.01\).
As for sampling, we empirically find that one-step ODE solving is sufficient for good performance in our problem, as validated in~\cref{sec:exp_ablation}.
\subsection{Baselines}
Our method is evaluated against multiple state-of-the-art approaches, including MotionCNN~\cite{konev2022motioncnn}, ReCoAt~\cite{huang2022recoat}, DenseTNT~\cite{gu2021densetnt}, SceneTF~\cite{ngiam2021scene}, HDGT~\cite{jia2023hdgt}, MTR~\cite{shi2022motion}, MTR++~\cite{shi2023mtr}, EDA~\cite{lin2024eda}, CtrlMTR~\cite{sun2024controlmtr}, and BeTop~\cite{liu2024reasoning}.
Their results are taken directly from the original papers and the official Waymo leaderboard to ensure fairness.

\subsection{Quantitative Results}
We present the main results in~\cref{tab:results_womd_motion} on the marginal prediction testing set of WOMD. Our method achieves competitive performance, outperforming all baselines across all metrics, including minADE, minFDE, Miss Rate, and two mAP metrics. It attains the highest mAP and soft mAP, and surpasses the other baselines by a non-trivial margin. The mAP and soft mAP metrics are the most crucial indicators of model performance according to the WOMD official evaluation protocol~\cite{ettinger2021large}.
Notably, these results were obtained using a single model for forecasting, without employing any ensemble techniques. To ensure a fair comparison, we exclude methods that utilize additional data, such as LIDAR or camera inputs.  
Please refer to~\cref{tab:more_results_womd} for results on the joint motion prediction dataset.

\subsection{Qualitative Results}
We present visualizations on the WOMD validation set to further assess our method. As shown in~\cref{fig:traj_success}, our model successfully predicts agent trajectories in complex traffic scenes, demonstrating its ability to capture intricate motion patterns. It effectively navigates challenging scenarios, such as turns, by maintaining a balance between accuracy and trajectory diversity. Notably, its confidence scores are well-calibrated, with high-confidence predictions consistently aligning with actual outcomes.

\begin{table}[!htbp]
\caption{\textbf{Ablation studies.}
We validate the effects of self-conditioning training and PL ranking loss.
}
\vspace{-0.1cm}
\centering
\scalebox{0.95}{
\begin{tabular}{c@{\hspace{6pt}}c@{\hspace{6pt}}c@{\hspace{6pt}}|c@{\hspace{6pt}}c@{\hspace{6pt}}c@{\hspace{6pt}}c}
\toprule
\begin{tabular}[l]{@{}>{}c@{}}Self-cond.\\Training\end{tabular} & \begin{tabular}[l]{@{}>{}c@{}}Ranking\\Loss\end{tabular} &  \begin{tabular}[l]{@{}>{}c@{}}ODE\\Steps\end{tabular} & minADE$\downarrow$ & minFDE$\downarrow$ & MissRate$\downarrow$ & mAP$\uparrow$ \\
\midrule
\multirow{3}{*}{\xmark} & \multirow{3}{*}{\cmark} & 1 & 0.6862 & 1.3967 & 0.1746 & 0.3669\\
 & & 5 & 0.7815 & 1.6348 & 0.2368 & 0.3007\\
 & & 10 & 0.9330 & 1.9749 & 0.2952 & 0.2398\\
\midrule
\multirow{3}{*}{\cmark} & \multirow{3}{*}{\cmark} & 1 & 0.6842 & 1.3858 & 0.1738 & 0.3695\\
 & & 5 & 0.6848 & 1.3849 & 0.1739 & 0.3659 \\
 & & 10 & 0.6839 & 1.3852 & 0.1736 & 0.3678 \\
 \midrule
\multirow{3}{*}{\cmark} & \multirow{3}{*}{\xmark} & 1 & 0.6901 & 1.4231 & 0.1764 & 0.3604 \\
 & & 5 & 0.6898 & 1.4237 & 0.1769 & 0.3608 \\
 & & 10 & 0.6903 & 1.4228 & 0.1785 & 0.3602 \\
\bottomrule
\end{tabular}
}
\vspace{-0.5cm}
\label{tab:exp_ablation}
\end{table}

\subsection{Ablation Studies}
\label{sec:exp_ablation}
To verify the introduced self-conditioning mechanism and the PL ranking loss, we conduct ablation studies by training various model variants using 20\% of the training set and evaluating them on the validation set. 
As shown in~\cref{tab:exp_ablation}, self-conditioning significantly improves performance as the number of ODE sampling steps increases (see~\cref{sect:method_prelim} for details on ODE sampling).  
In contrast, without self-conditioning, the model's generalization degrades with more sampling steps due to the overfitting issue discussed in~\cref{sect:method_overfit}.  
Additionally, we compare our approach with a model variant that does not incorporate the PL ranking loss, as defined in~\cref{eq:loss_rank}. The results clearly demonstrate that the learning-to-rank regularization loss effectively enhances the model's performance across a wide range of metrics, particularly in mAP.
Because mAP effectively measures the alignment between trajectory scores and prediction accuracy, its improvement with the PL ranking loss indicates a meaningful enhancement in uncertainty estimation.

\section{Conclusion}
In this paper, we present \modelName, a flow-matching-based model for multi-modal motion prediction.  
Our model enables generating multiple future trajectories with confidence scores in a single sampling pass, eliminating the need for repetitive \iid sampling required by conventional generative models.  
Additionally, we introduce a principled method to improve confidence score calibration via the PL ranking loss.
Our model achieves state-of-the-art performance on the large-scale real-world WOMD dataset, providing a novel and efficient solution to motion prediction.


\section*{Acknowledgment}
This work was funded, in part, by the NSERC DG Grant (No. RGPIN-2022-04636), the Vector Institute for AI, Canada CIFAR AI Chair, and a XMotors.ai Gift Fund. 
Resources used in preparing this research were provided, in part, by the Province of Ontario, the Government of Canada through the Digital Research Alliance of Canada \url{alliance.can.ca}, and companies sponsoring the Vector Institute \url{www.vectorinstitute.ai/#partners}, and Advanced Research Computing at the University of British Columbia. 
Additional hardware support was provided by John R. Evans Leaders Fund CFI grant.
QY is supported by UBC Four Year Doctoral Fellowships.



%



\bibliographystyle{IEEEtran}
\bibliography{ref}

\begin{thebibliography}{10}
\providecommand{\url}[1]{#1}
\csname url@samestyle\endcsname
\providecommand{\newblock}{\relax}
\providecommand{\bibinfo}[2]{#2}
\providecommand{\BIBentrySTDinterwordspacing}{\spaceskip=0pt\relax}
\providecommand{\BIBentryALTinterwordstretchfactor}{4}
\providecommand{\BIBentryALTinterwordspacing}{\spaceskip=\fontdimen2\font plus
\BIBentryALTinterwordstretchfactor\fontdimen3\font minus \fontdimen4\font\relax}
\providecommand{\BIBforeignlanguage}[2]{{%
\expandafter\ifx\csname l@#1\endcsname\relax
\typeout{** WARNING: IEEEtran.bst: No hyphenation pattern has been}%
\typeout{** loaded for the language `#1'. Using the pattern for}%
\typeout{** the default language instead.}%
\else
\language=\csname l@#1\endcsname
\fi
#2}}
\providecommand{\BIBdecl}{\relax}
\BIBdecl

\bibitem{liang2020learning}
M.~Liang, B.~Yang, R.~Hu, Y.~Chen, R.~Liao, S.~Feng, and R.~Urtasun, ``Learning lane graph representations for motion forecasting,'' in \emph{ECCV}, 2020.

\bibitem{casas2020implicit}
S.~Casas, C.~Gulino, S.~Suo, K.~Luo, R.~Liao, and R.~Urtasun, ``Implicit latent variable model for scene-consistent motion forecasting,'' in \emph{ECCV}, 2020.

\bibitem{luo2021safety}
K.~Luo, S.~Casas, R.~Liao, X.~Yan, Y.~Xiong, W.~Zeng, and R.~Urtasun, ``Safety-oriented pedestrian occupancy forecasting,'' in \emph{IROS}, 2021.

\bibitem{zeng2021lanercnn}
W.~Zeng, M.~Liang, R.~Liao, and R.~Urtasun, ``Lanercnn: Distributed representations for graph-centric motion forecasting,'' in \emph{IROS}, 2021.

\bibitem{konev2022motioncnn}
S.~Konev, K.~Brodt, and A.~Sanakoyeu, ``Motioncnn: a strong baseline for motion prediction in autonomous driving,'' \emph{arXiv preprint arXiv:2206.02163}, 2022.

\bibitem{liu2021social}
Y.~Liu, Q.~Yan, and A.~Alahi, ``Social nce: Contrastive learning of socially-aware motion representations,'' in \emph{ICCV}, 2021.

\bibitem{chang2019argoverse}
M.-F. Chang, J.~Lambert, P.~Sangkloy, J.~Singh, S.~Bak, A.~Hartnett, D.~Wang, P.~Carr, S.~Lucey, D.~Ramanan \emph{et~al.}, ``Argoverse: 3d tracking and forecasting with rich maps,'' in \emph{CVPR}, 2019.

\bibitem{ettinger2021large}
S.~Ettinger, S.~Cheng, B.~Caine, C.~Liu, H.~Zhao, S.~Pradhan, Y.~Chai, B.~Sapp, C.~R. Qi, Y.~Zhou \emph{et~al.}, ``Large scale interactive motion forecasting for autonomous driving: The waymo open motion dataset,'' in \emph{ICCV}, 2021.

\bibitem{zhao2021tnt}
H.~Zhao, J.~Gao, T.~Lan, C.~Sun, B.~Sapp, B.~Varadarajan, Y.~Shen, Y.~Shen, Y.~Chai, C.~Schmid \emph{et~al.}, ``Tnt: Target-driven trajectory prediction,'' in \emph{CoRL}, 2021.

\bibitem{gu2021densetnt}
J.~Gu, C.~Sun, and H.~Zhao, ``Densetnt: End-to-end trajectory prediction from dense goal sets,'' in \emph{ICCV}, 2021.

\bibitem{Gu_2022_CVPR}
T.~Gu, G.~Chen, J.~Li, C.~Lin, Y.~Rao, J.~Zhou, and J.~Lu, ``Stochastic trajectory prediction via motion indeterminacy diffusion,'' in \emph{CVPR}, 2022.

\bibitem{Mao_2023_CVPR}
W.~Mao, C.~Xu, Q.~Zhu, S.~Chen, and Y.~Wang, ``Leapfrog diffusion model for stochastic trajectory prediction,'' in \emph{CVPR}, 2023.

\bibitem{jiang2023motiondiffuser}
C.~Jiang, A.~Cornman, C.~Park, B.~Sapp, Y.~Zhou, D.~Anguelov \emph{et~al.}, ``Motiondiffuser: Controllable multi-agent motion prediction using diffusion,'' in \emph{CVPR}, 2023.

\bibitem{fu2025moflow}
Y.~Fu, Q.~Yan, L.~Wang, K.~Li, and R.~Liao, ``Moflow: One-step flow matching for human trajectory forecasting via implicit maximum likelihood estimation based distillation,'' \emph{CVPR}, 2025.

\bibitem{seff2023motionlm}
A.~Seff, B.~Cera, D.~Chen, M.~Ng, A.~Zhou, N.~Nayakanti, K.~S. Refaat, R.~Al-Rfou, and B.~Sapp, ``Motionlm: Multi-agent motion forecasting as language modeling,'' in \emph{ICCV}, 2023.

\bibitem{ngiam2021scene}
J.~Ngiam, B.~Caine, V.~Vasudevan, Z.~Zhang, H.-T.~L. Chiang, J.~Ling, R.~Roelofs, A.~Bewley, C.~Liu, A.~Venugopal \emph{et~al.}, ``Scene transformer: A unified architecture for predicting multiple agent trajectories,'' \emph{arXiv preprint arXiv:2106.08417}, 2021.

\bibitem{nayakanti2023wayformer}
N.~Nayakanti, R.~Al-Rfou, A.~Zhou, K.~Goel, K.~S. Refaat, and B.~Sapp, ``Wayformer: Motion forecasting via simple \& efficient attention networks,'' in \emph{ICRA}, 2023.

\bibitem{varadarajan2022multipath++}
B.~Varadarajan, A.~Hefny, A.~Srivastava, K.~S. Refaat, N.~Nayakanti, A.~Cornman, K.~Chen, B.~Douillard, C.~P. Lam, D.~Anguelov \emph{et~al.}, ``Multipath++: Efficient information fusion and trajectory aggregation for behavior prediction,'' in \emph{ICRA}, 2022.

\bibitem{shi2022motion}
S.~Shi, L.~Jiang, D.~Dai, and B.~Schiele, ``Motion transformer with global intention localization and local movement refinement,'' \emph{NeurIPS}, 2022.

\bibitem{shi2023mtr}
------, ``Mtr++: Multi-agent motion prediction with symmetric scene modeling and guided intention querying,'' \emph{IEEE TPAMI}, 2024.

\bibitem{zhou2023query}
Z.~Zhou, J.~Wang, Y.-H. Li, and Y.-K. Huang, ``Query-centric trajectory prediction,'' in \emph{CVPR}, 2023.

\bibitem{jia2023hdgt}
X.~Jia, P.~Wu, L.~Chen, Y.~Liu, H.~Li, and J.~Yan, ``Hdgt: Heterogeneous driving graph transformer for multi-agent trajectory prediction via scene encoding,'' \emph{IEEE TPAMI}, 2023.

\bibitem{lin2024eda}
L.~Lin, X.~Lin, T.~Lin, L.~Huang, R.~Xiong, and Y.~Wang, ``Eda: Evolving and distinct anchors for multimodal motion prediction,'' in \emph{AAAI}, 2024.

\bibitem{liu2024reasoning}
H.~Liu, L.~Chen, Y.~Qiao, C.~Lv, and H.~Li, ``Reasoning multi-agent behavioral topology for interactive autonomous driving,'' \emph{NeurIPS}, vol.~37, 2024.

\bibitem{gupta2018social}
A.~Gupta, J.~Johnson, L.~Fei-Fei, S.~Savarese, and A.~Alahi, ``Social gan: Socially acceptable trajectories with generative adversarial networks,'' in \emph{CVPR}, 2018.

\bibitem{dendorfer2021mg}
P.~Dendorfer, S.~Elflein, and L.~Leal-Taix{\'e}, ``Mg-gan: A multi-generator model preventing out-of-distribution samples in pedestrian trajectory prediction,'' in \emph{ICCV}, 2021.

\bibitem{groupnet_CVPR}
C.~Xu, M.~Li, Z.~Ni, Y.~Zhang, and S.~Chen, ``Groupnet: Multiscale hypergraph neural networks for trajectory prediction with relational reasoning,'' in \emph{CVPR}, 2022.

\bibitem{xu2024dynamic}
C.~Xu, Y.~Wei, B.~Tang, S.~Yin, Y.~Zhang, S.~Chen, and Y.~Wang, ``Dynamic-group-aware networks for multi-agent trajectory prediction with relational reasoning,'' \emph{Neural Networks}, vol. 170, 2024.

\bibitem{Lee_2022_CVPR}
M.~Lee, S.~S. Sohn, S.~Moon, S.~Yoon, M.~Kapadia, and V.~Pavlovic, ``Muse-vae: Multi-scale vae for environment-aware long term trajectory prediction,'' in \emph{CVPR}, 2022.

\bibitem{lipman2022flow}
Y.~Lipman, R.~T. Chen, H.~Ben-Hamu, M.~Nickel, and M.~Le, ``Flow matching for generative modeling,'' \emph{ICLR}, 2023.

\bibitem{vaswani2017attention}
A.~Vaswani, N.~Shazeer, N.~Parmar, J.~Uszkoreit, L.~Jones, A.~N. Gomez, {\L}.~Kaiser, and I.~Polosukhin, ``Attention is all you need,'' \emph{NeurIPS}, 2017.

\bibitem{ho2020denoising}
J.~Ho, A.~Jain, and P.~Abbeel, ``Denoising diffusion probabilistic models,'' \emph{NeurIPS}, vol.~33, 2020.

\bibitem{chen2023executing}
X.~Chen, B.~Jiang, W.~Liu, Z.~Huang, B.~Fu, T.~Chen, and G.~Yu, ``Executing your commands via motion diffusion in latent space,'' in \emph{Proceedings of the IEEE/CVF conference on computer vision and pattern recognition}, 2023, pp. 18\,000--18\,010.

\bibitem{tevet2022human}
G.~Tevet, S.~Raab, B.~Gordon, Y.~Shafir, D.~Cohen-Or, and A.~H. Bermano, ``Human motion diffusion model,'' \emph{arXiv preprint arXiv:2209.14916}, 2022.

\bibitem{zhang2024motiondiffuse}
M.~Zhang, Z.~Cai, L.~Pan, F.~Hong, X.~Guo, L.~Yang, and Z.~Liu, ``Motiondiffuse: Text-driven human motion generation with diffusion model,'' \emph{IEEE transactions on pattern analysis and machine intelligence}, vol.~46, no.~6, pp. 4115--4128, 2024.

\bibitem{liu2024geneoh}
X.~Liu and L.~Yi, ``Geneoh diffusion: Towards generalizable hand-object interaction denoising via denoising diffusion,'' \emph{arXiv preprint arXiv:2402.14810}, 2024.

\bibitem{lee2024interhandgen}
J.~Lee, S.~Saito, G.~Nam, M.~Sung, and T.-K. Kim, ``Interhandgen: Two-hand interaction generation via cascaded reverse diffusion,'' in \emph{Proceedings of the IEEE/CVF Conference on Computer Vision and Pattern Recognition}, 2024, pp. 527--537.

\bibitem{Li_2025_CVPR}
M.~Li, S.~Christen, C.~Wan, Y.~Cai, R.~Liao, L.~Sigal, and S.~Ma, ``Latenthoi: On the generalizable hand object motion generation with latent hand diffusion.'' in \emph{Proceedings of the Computer Vision and Pattern Recognition Conference (CVPR)}, June 2025, pp. 17\,416--17\,425.

\bibitem{qi2017pointnet}
C.~R. Qi, H.~Su, K.~Mo, and L.~J. Guibas, ``Pointnet: Deep learning on point sets for 3d classification and segmentation,'' in \emph{CVPR}, 2017.

\bibitem{plackett}
R.~L. Plackett, ``The analysis of permutations,'' \emph{Applied Statistics}, vol.~24, no.~2, 1975.

\bibitem{luce}
R.~D. Luce, \emph{Individual Choice Behavior}.\hskip 1em plus 0.5em minus 0.4em\relax John Wiley, 1959.

\bibitem{zhang2024symmetricdiffusers}
Y.~Zhang, D.~Yang, and R.~Liao, ``Symmetricdiffusers: Learning discrete diffusion on finite symmetric groups,'' \emph{ICLR}, 2025.

\bibitem{huang2022recoat}
Z.~Huang, X.~Mo, and C.~Lv, ``Recoat: A deep learning-based framework for multi-modal motion prediction in autonomous driving application,'' in \emph{IEEE 25th International Conference on Intelligent Transportation Systems (ITSC)}, 2022.

\bibitem{sun2024controlmtr}
J.~Sun, C.~Yuan, S.~Sun, S.~Wang, Y.~Han, S.~Ma, Z.~Huang, A.~Wong, K.~P. Tee, and M.~H. Ang~Jr, ``Controlmtr: Control-guided motion transformer with scene-compliant intention points for feasible motion prediction,'' in \emph{IEEE 27th International Conference on Intelligent Transportation Systems (ITSC)}, 2024.

\bibitem{loshchilov2017decoupled}
I.~Loshchilov and F.~Hutter, ``Decoupled weight decay regularization,'' \emph{ICLR}, 2019.

\bibitem{paszke2019pytorch}
A.~Paszke, S.~Gross, F.~Massa, A.~Lerer, J.~Bradbury, G.~Chanan, T.~Killeen, Z.~Lin, N.~Gimelshein, L.~Antiga \emph{et~al.}, ``Pytorch: An imperative style, high-performance deep learning library,'' \emph{NeurIPS}, 2019.

\bibitem{chen2022analog}
T.~Chen, R.~Zhang, and G.~Hinton, ``Analog bits: Generating discrete data using diffusion models with self-conditioning,'' \emph{arXiv preprint arXiv:2208.04202}, 2022.

\bibitem{yan2023swingnn}
Q.~Yan, Z.~Liang, Y.~Song, R.~Liao, and L.~Wang, ``Swingnn: Rethinking permutation invariance in diffusion models for graph generation,'' \emph{arXiv preprint arXiv:2307.01646}, 2023.

\bibitem{xu2024joint}
B.~Xu, Q.~Yan, R.~Liao, L.~Wang, and L.~Sigal, ``Joint generative modeling of scene graphs and images via diffusion models,'' \emph{arXiv preprint arXiv:2401.01130}, 2024.

\bibitem{jiao2025uniedit}
G.~Jiao, B.~Huang, K.-C. Wang, and R.~Liao, ``Uniedit-flow: Unleashing inversion and editing in the era of flow models,'' \emph{arXiv preprint arXiv:2504.13109}, 2025.

\bibitem{yang2025qdm}
D.~Yang, P.~Vicol, X.~Qi, R.~Liao, and X.~Zhang, ``Qdm: Quadtree-based region-adaptive sparse diffusion models for efficient image super-resolution,'' \emph{arXiv preprint arXiv:2503.12015}, 2025.

\bibitem{song2021scorebased}
\BIBentryALTinterwordspacing
Y.~Song, J.~Sohl-Dickstein, D.~P. Kingma, A.~Kumar, S.~Ermon, and B.~Poole, ``Score-based generative modeling through stochastic differential equations,'' in \emph{ICLR}, 2021. [Online]. Available: \url{https://openreview.net/forum?id=PxTIG12RRHS}
\BIBentrySTDinterwordspacing

\bibitem{Karras2022edm}
T.~Karras, M.~Aittala, T.~Aila, and S.~Laine, ``Elucidating the design space of diffusion-based generative models,'' in \emph{Proc. NeurIPS}, 2022.

\bibitem{melnik2024video}
A.~Melnik, M.~Ljubljanac, C.~Lu, Q.~Yan, W.~Ren, and H.~Ritter, ``Video diffusion models: A survey,'' \emph{arXiv preprint arXiv:2405.03150}, 2024.

\bibitem{mo2021multi}
X.~Mo, Z.~Huang, and C.~Lv, ``Multi-modal interactive agent trajectory prediction using heterogeneous edge-enhanced graph attention network,'' in \emph{Workshop on Autonomous Driving, CVPR}, vol.~6, 2021, p.~7.

\bibitem{wu2021air}
D.~Wu and Y.~Wu, ``$\text{AIR}^2$ for interaction prediction,'' \emph{arXiv preprint arXiv:2111.08184}, 2021.

\bibitem{sun2022m2i}
Q.~Sun, X.~Huang, J.~Gu, B.~C. Williams, and H.~Zhao, ``M2i: From factored marginal trajectory prediction to interactive prediction,'' in \emph{Proceedings of the IEEE/CVF Conference on Computer Vision and Pattern Recognition}, 2022, pp. 6543--6552.

\bibitem{huang2023gameformer}
Z.~Huang, H.~Liu, and C.~Lv, ``Gameformer: Game-theoretic modeling and learning of transformer-based interactive prediction and planning for autonomous driving,'' in \emph{Proceedings of the IEEE/CVF International Conference on Computer Vision}, 2023, pp. 3903--3913.

\bibitem{jia2024amp}
X.~Jia, S.~Shi, Z.~Chen, L.~Jiang, W.~Liao, T.~He, and J.~Yan, ``Amp: Autoregressive motion prediction revisited with next token prediction for autonomous driving,'' \emph{arXiv preprint arXiv:2403.13331}, 2024.

\end{thebibliography}

\clearpage
\appendix
\section{Appendix}

\subsection{Implementation Details}
\subsubsection{Training and Inference Pipeline}
\begin{figure}[htbp]
    \vspace{-0.75cm}
    \begin{minipage}[t]{0.48\textwidth} 
    \begin{algorithm}[H] 
    \caption{\modelName Training}   
    \label{alg:train}
    \begin{algorithmic}[1]
        \State\textbf{Require:}\xspace Training set $p_\mathcal{D}$, learning rate $\eta$, model $F_\theta$
        \Repeat 
            \State $(Y^1, C) \sim p_\mathcal{D}$ \Comment{Sample motion and context data}
            \State $Y^0 \sim \mathcal{N}(\b 0, \b I)$
            \State $t \sim p_t(t)$ \Comment{Flow matching time}
            \State $Y^t \leftarrow t Y^1 + (1-t) Y^0$
            \State $p_s \sim U(0, 1)$
            \If{$p_s < 0.5$} \Comment{Self-Conditioning}
                \State $\check{Y}^1, \check{S}, \check{r} \leftarrow F_\theta(Y^t, C, t)$
                \State $Y^t \leftarrow t \check{Y}^1 + (1-t) Y^0$
                \State $l_s \gets L(\check{S}, \check{Y}^1, \check{r}, Y^1)$ \Comment{Loss in \cref{eq:loss_fm_total}}
            \Else
                \State $l_s \gets 0$
            \EndIf
            \State $\hat{Y}^1, S, r \leftarrow F_\theta(Y^t, C, t)$
            \State $l \gets L(S, \hat{Y}^1, r, Y^1)$ \Comment{Loss in \cref{eq:loss_fm_total}}
            \State $\theta\gets\theta-\eta\nabla_{\theta} (l +l_s)$
        \Until{converged}
        \end{algorithmic}
    \end{algorithm}
    \end{minipage}\hfill
    \begin{minipage}[t]{0.48\textwidth} 
        \begin{algorithm}[H] 
            \caption{\modelName Sampling}   
            \label{alg:sample}
            \begin{algorithmic}[1]
                \State \textbf{Require:} Testing set \(p_\mathcal{D}'\), model \(F_\theta\), sampling steps \(T\)
                \Repeat 
                    \State \((\cdot, C) \sim p_\mathcal{D}'\) \Comment{Sample context data}
                    \State \(Y^0 \sim \mathcal{N}(\mathbf{0}, \mathbf{I})\)
                    \For {\(n \in [0, 1, \cdots, T-1]\)}
                        \State $t \leftarrow n/T$
                        \State \(\hat{Y}^1, S, r \gets F_\theta(Y^t, C, t)\)
                        \State \(v_\theta \gets \frac{\hat{Y}^1 - Y^t}{1 - t}\)
                        \State \(Y^{t + 1/T} \gets Y^{t} + \frac{1}{T} v_\theta\)
                    \EndFor
                \Until{finished}
            \end{algorithmic}
        \end{algorithm}
    \end{minipage}
\end{figure}
The pseudo-code for the training and sampling procedures utilized in the \modelName framework is provided in \cref{alg:train,alg:sample}.
During training, a composite objective is employed, integrating flow-matching denoising loss, classification loss, and a learning-to-rank loss to encourage diverse and accurate predictions. Additionally, self-conditioning is incorporated to mitigate overfitting in conditional generative models, particularly when the noise injection parameter $t$ is small.
The self-conditioning technique used here shares similarity with that in~\cite{chen2022analog}; however, we use the initial denoised result to modify the noise-corrupted input data rather than concatenating to augment the input. With a 50\% probability of applying self-conditioning during training, this technique adds merely a small amount of computational cost to the training pipeline in memory use and runtime. In practice, the optimizations in lines 11 and 14 of~\cref{alg:train} can often be merged into a single operation. The context encoder's forward pass can also use a caching mechanism to accelerate computation. As for the sampling procedure, we follow the standard flow-matching framework, operating along ODE trajectories as shown in~\cref{alg:sample}. Note that there is no need to implement the self-conditioning step in the sampling process.

\subsubsection{Model Configurations}
\begin{table}[htbp]
\centering
\caption{\textbf{Detailed model configuration of {\modelName}.}}
\label{tab:model_config}
\renewcommand{\arraystretch}{1.2}%
\begin{tabular}{>{\bfseries}p{0.42\linewidth} p{0.48\linewidth}}
\toprule
\rowcolor[HTML]{EFEFEF} 
\textbf{Parameter} & \textbf{Value} \\
\midrule
\multicolumn{2}{l}{\textbf{Context Encoder}} \\
\quad layers & 6 \\
\quad embed dim & 256 \\
\quad attention heads & 8 (self-attention)\\
\midrule
\multicolumn{2}{l}{\textbf{Flow Matching Decoder}} \\
\quad layers & 6 \\
\quad embed dim & 512 \\
\quad attention heads & 8 (cross-attention) \\
\midrule
\multicolumn{2}{l}{\textbf{Prediction Heads}} \\
\quad trajectory head & 3-layer MLP \\
\quad classification head & 3-layer MLP \\
\quad ranking head & 3-layer MLP \\
\midrule
\multicolumn{2}{l}{\textbf{Training}} \\
\quad optimizer & AdamW \\
\quad Adam $(\beta_1, \beta_2)$ & $(0.9, 0.999)$ \\
\quad lr scheduler & Linear Decay \\
\quad epochs & 40 \\
\quad batch size & 80 \\
\quad peak learning rate & $1 \times 10^{-4}$ \\
\quad weight decay & 0.01 \\
\midrule
\multicolumn{2}{l}{\textbf{Loss Weights}} \\
\quad $L_{\text{flow}}$ & 1.0 \\
\quad $L_{\text{cls}}$ & 1.0 \\
\quad $L_{\text{rank}}$ & 0.1 \\
\bottomrule
\end{tabular}
\end{table}

As shown in Table~\ref{tab:model_config}, \textit{\modelName} primarily comprises a context encoder, flow matching decoder, and prediction heads. The context encoder has 6 layers, an embedding dimension of 256, and 8 attention heads. The flow matching decoder also has 6 layers but with a larger embedding dimension of 512 and 8 attention heads. The prediction heads for trajectory, classification, and ranking each use a 3-layer MLP. Training is conducted using the AdamW optimizer~\cite{loshchilov2017decoupled} with $\beta_1=0.9$, $\beta_2=0.999$, a linear decay learning rate scheduler over 22/24/26/28 epochs, 40 total epochs, a batch size of 80, a peak learning rate of $1 \times 10^{-4}$, and a weight decay of 0.01. Loss weights for flow matching denoising ($L_{\text{flow}}$) and classification ($L_{\text{cls}}$) are set to 1.0 and the weight for ranking ($L_{\text{rank}}$) is set to 0.1.
To adapt to the denoising objective, we map trajectory data (X-Y coordinates) to the range $[-1, 1]$ using linear-based normalization. For each coordinate, we subtract an offset and multiply by a scaling coefficient to ensure consistent scaling across the dataset based on predefined metadata parameters. We collect statistics from the training set to ensure that over 99.9\% of data points can be reversibly mapped to the target range without loss. This normalization aligns with previous work on diffusion and flow-matching generative models~\cite{yan2023swingnn, xu2024joint, jiao2025uniedit, yang2025qdm, song2021scorebased, lipman2022flow, Karras2022edm, melnik2024video}, which aims to improve training stability.

\begin{table*}[htbp]
\centering
\caption{\textbf{Additional results on WOMD dataset.} For a fair comparison, we compare our method with existing motion prediction approaches without using model ensembling or extra data such as LIDAR.
}
\label{tab:more_results_womd}
\begin{tabular}{l|l|ccc|cc}
\toprule
Set split & Method & minADE $\downarrow$ & minFDE $\downarrow$ & Miss Rate $\downarrow$ & mAP $\uparrow$  & {Soft mAP} $\uparrow$ \\ \midrule
\multirow{11}{*}{\texttt{Standard-Test}} & MotionCNN~\cite{konev2022motioncnn} & 0.7400 & 1.4936 & 0.2091 & 0.2136 & - \\
                      & ReCoAt~\cite{huang2022recoat} & 0.7703 & 1.6668  & 0.2437 & 0.2711 & - \\
                      & DenseTNT~\cite{gu2021densetnt} & 1.0387 & 1.5514 & 0.1573 & 0.3281 & - \\
                      & SceneTransformer~\cite{ngiam2021scene} & 0.6117 & 1.2116 & 0.1564 & 0.2788 & - \\
                      & HDGT~\cite{jia2023hdgt}     & 0.5933 & 1.2055 & 0.1854& 0.3577 & 0.3709 \\
                      & MTR~\cite{shi2022motion}     & 0.6050 & 1.2207 & 0.1351 & 0.4129 & 0.4216 \\ 
                      & MTR++~\cite{shi2023mtr}     & 0.5906 & 1.1939 & 0.1298  & 0.4329 & 0.4414 \\  
                      & EDA~\cite{lin2024eda}  & \underline{0.5718} & {1.1702}  & \underline{0.1169}  & 
                      {0.4487} & {0.4596} \\ 
                      & ControlMTR~\cite{sun2024controlmtr}   & 0.5897 & 1.1916 & 0.1282  & 0.4414 & 0.4572 \\ 
                      & {BeTop}~\cite{liu2024reasoning}     & {0.5723} & \underline{1.1668} & {0.1176} & \underline{0.4566} & \underline{0.4678} \\
                      & \textbf{\modelName} & \textbf{0.5714} & \textbf{1.1667} & \textbf{0.1162} & \textbf{0.4604} & \textbf{0.4710} \\ 
                      \midrule
\multirow{4}{*}{\texttt{Standard-Val}} & MTR~\cite{shi2022motion}     & 0.6046 & 1.2251 & 0.1366  & 0.4164 &  - \\ 
                     & EDA~\cite{lin2024eda}     & \textbf{0.5708} & \underline{1.1730} & \underline{0.1178}  & {0.4353} & \underline{0.4462} \\ 
                     & {BeTop}~\cite{liu2024reasoning}     & \underline{0.5716} & \textbf{1.1640} & \textbf{0.1177} & \underline{0.4416}  & -\\ 
                     & \textbf{\modelName} & {0.5734} & {1.1768} & {0.1187} & \textbf{0.4539} & \textbf{0.4646} \\ 
                     \midrule
\multirow{10}{*}{\texttt{Inter.-Test}} & HeatIRm4~\cite{mo2021multi} & 1.4197 & 3.2595  & 0.7224 & 0.0804 & - \\
                      & AIR\textsuperscript{2}~\cite{wu2021air} & 1.3165 & 2.7138 & 0.6230 & 0.0963 & - \\
                      & SceneTransformer~\cite{ngiam2021scene} & 0.9774 & 2.1892 & 0.4942 & 0.1192 & - \\
                      & M2I~\cite{sun2022m2i}    & 1.3506 & 2.8325 & 0.5538 &0.1239 & -  \\
                      & GameFormer~\cite{huang2023gameformer}  & 0.9721 & 2.2146 & 0.4933 & 0.1923 & 0.1982 \\
                      & AMP~\cite{jia2024amp}     & \underline{0.9073} & \underline{2.0415} & {0.4212} & 0.2294 & 0.2365 \\ 
                      & MTR~\cite{shi2022motion} & 0.9181 & 2.0633 & 0.4411 & 0.2037 & - \\
                      & MTR++~\cite{shi2023mtr}     & \textbf{0.8795} & \textbf{1.9505} & \textbf{0.4143}  & {0.2326} & {0.2368}  \\  
                      & BeTop~\cite{liu2024reasoning}  & 0.9744 & 2.2744 & 0.4355 & \underline{0.2412} & \underline{0.2466} \\
                      & \textbf{\modelName (ours)}  & 0.9381 & 2.1521 & \underline{0.4180} & \textbf{0.2533} & \textbf{0.2593} \\ 
                      \midrule
\multirow{3}{*}{\texttt{Inter.-Val}} & MTR~\cite{shi2022motion}    & \underline{0.9132} & \underline{2.0536} & 0.4372 & 0.1992 &- \\ 
                     & AMP~\cite{jia2024amp}    & \textbf{0.8910} & \textbf{2.0133} & {0.4172} & {0.2344} &- \\ 
                     & BeTop~\cite{liu2024reasoning} & 0.9304 & 2.1340 & \underline{0.4154} & \underline{0.2366} &-  \\ 
                     & \textbf{\modelName (Ours)} & 0.9273 & 2.1249 & \textbf{0.4110} & \textbf{0.2464} & \textbf{0.2531}  \\ 
                      \bottomrule
\end{tabular}
\vspace{-12pt}
\end{table*}
\subsection{Additional Experimental Results}
\subsubsection{More Quantitative Results}
We present additional quantitative results in~\cref{tab:more_results_womd}, covering both the standard validation and test splits, as well as the interactive validation and test sets.
The WOMD benchmark includes two tasks, each with its own evaluation metrics: (1) the standard (marginal) motion prediction task, which independently assesses the predicted trajectories of up to 8 agents per scene; and (2) the interactive (joint) motion prediction task, which evaluates the joint future trajectories of 2 interacting agents. Both tasks provide 1 second of historical data and require predicting 6 candidate marginal or joint trajectories for 8 seconds into the future.
To evaluate our approach to interactive dataset, we combine the marginal predictions of two interacting agents into joint predictions following the method in~\cite{shi2022motion, shi2023mtr}. We generate 36 (6\textsuperscript{2}) joint trajectory combinations from the individual predictions and select the top 6 based on their combined confidence, computed as the product of the agents' marginal probabilities.
Please note that due to changes in the Waymo official evaluation server and internal protocols, there may be slight mismatches in the baseline results (which were evaluated using the available protocol at the time of their publication). We aim to offer fair comparisons as much as possible within our available capacity.

Our method outperforms all baselines across all metrics, including ADE, FDE, Miss Rate, and mAP, on the standard test split of the dataset.
On the standard validation set, our method performs strongly in terms of mAP and Soft mAP, indicating a robust capacity for modeling uncertainty, while maintaining a very small gap compared to the state-of-the-art in ADE, FDE, and Miss Rate.
On the interactive dataset, our method surpasses the baselines in both mAP and Soft mAP.
Please note that, due to the lack of open-source resources, we were unable to fully reproduce the results of previous models on the interactive sets; therefore, we compare only against the numbers reported in their respective papers.

\subsubsection{Runtime Comparison}
\begin{table}[h]
\centering
\caption{\textbf{Inference runtime comparison.}}
\begin{tabular}{c|cc|c}
\hline
\multirow{2}{*}{{Method}} & \multicolumn{2}{c|}{{Runtime}} & \multirow{2}{*}{{\#Params}} \\
                                 & {Per Agent (ms)$\downarrow$} & {Total Val. Set (s)$\downarrow$} & \\
\hline
MTR~\cite{shi2022motion}   & 19.96 & 880 & 65.8M \\
Ours  & 16.44 & 725 & 53.3M \\
\hline
\end{tabular}
\label{tab:runtime}
\end{table}
We present the inference runtime results in~\cref{tab:runtime}, comparing MTR~\cite{shi2022motion} and our method on the WOMD validation set, which consists of 44,097 agents, using the same hardware (NVIDIA A100 80GB).
Thanks to the multi-shot prediction formulation and few-step sampling, our method achieves relatively fast runtime.
Also, our model is slightly smaller than MTR, which further contributes to its faster inference speed.

\begin{figure*}[htbp]
    \centering
    \includegraphics[width=0.48\textwidth]{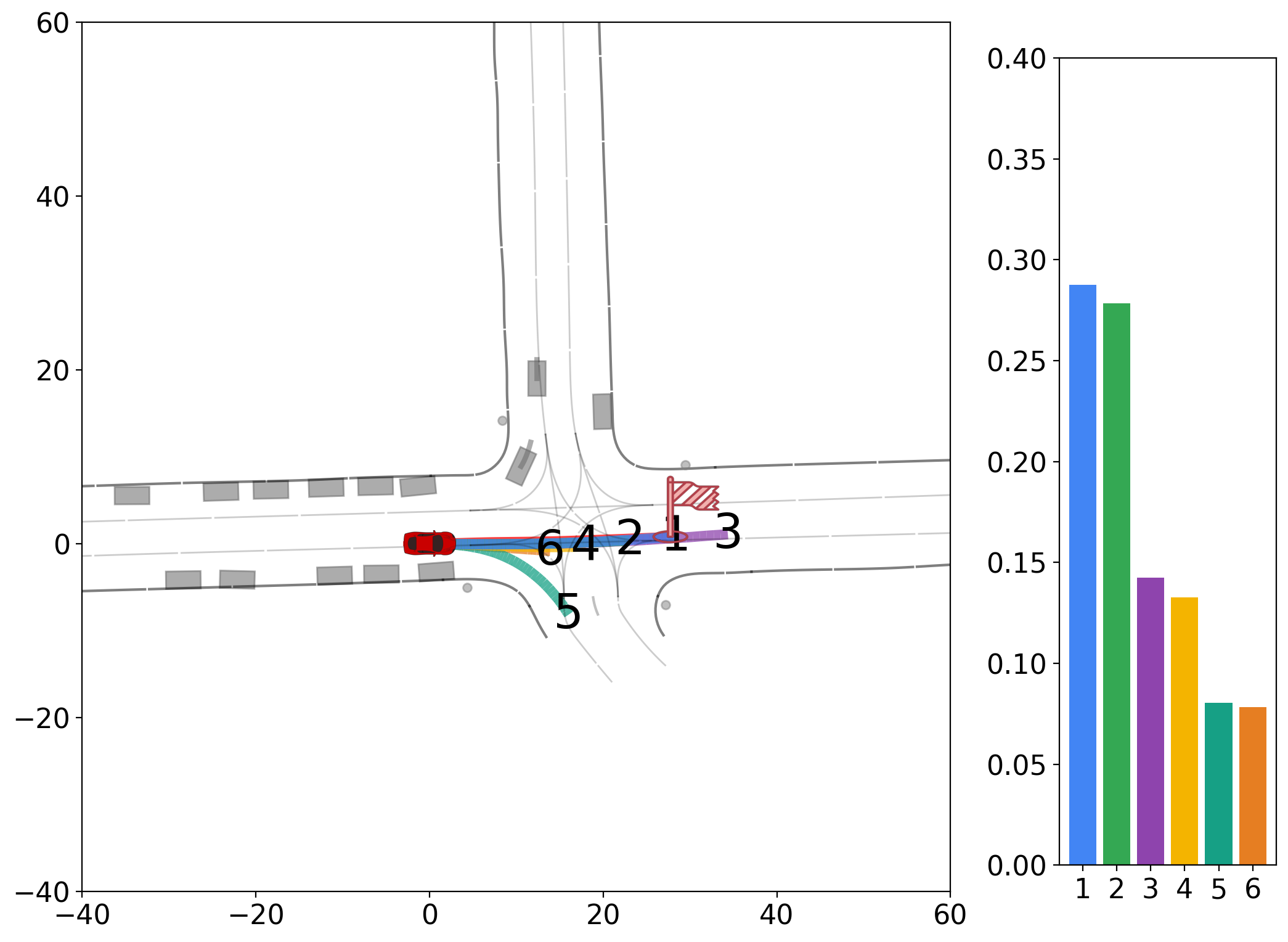} 
    \includegraphics[width=0.48\textwidth]{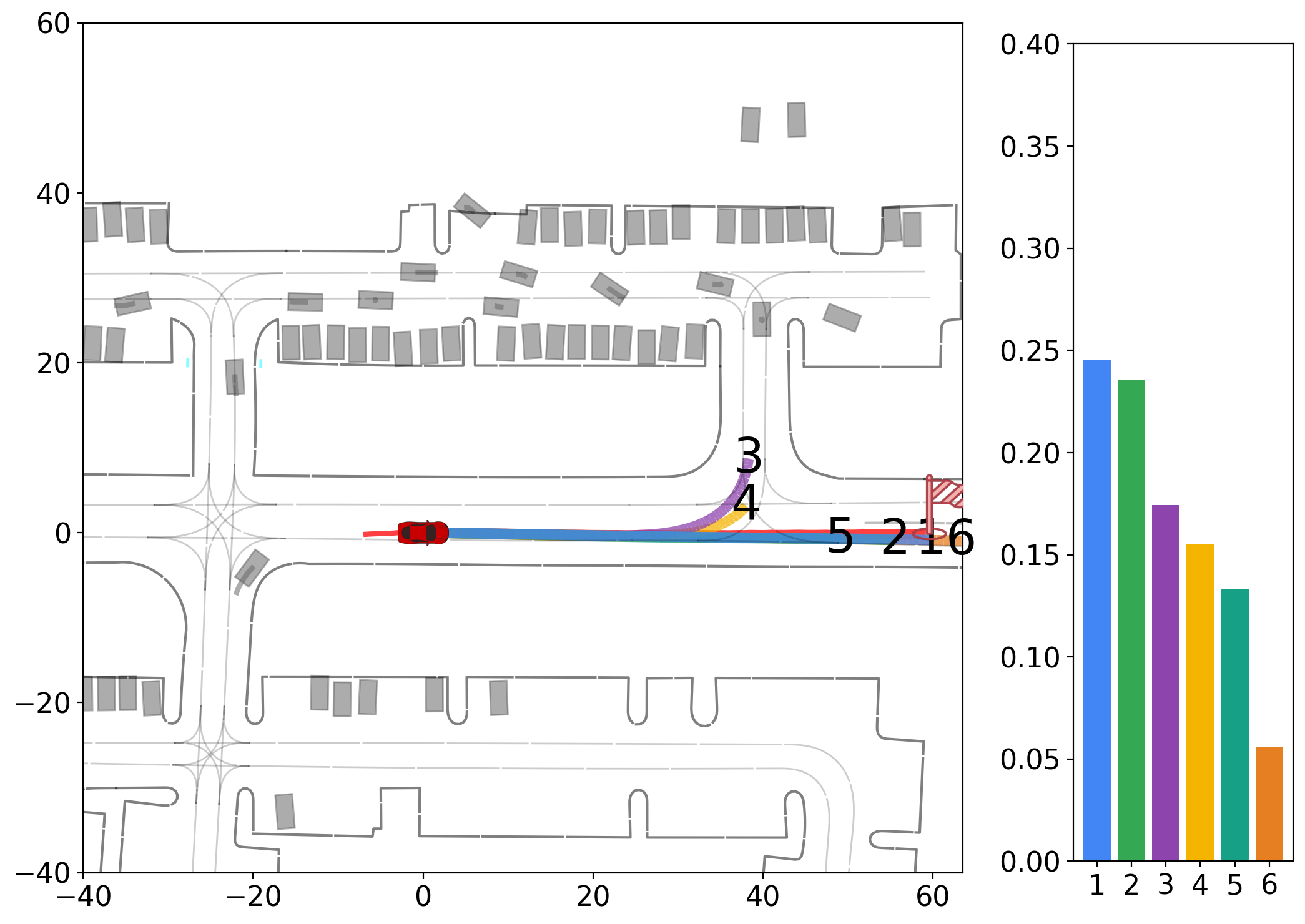}
    \includegraphics[width=0.48\textwidth]{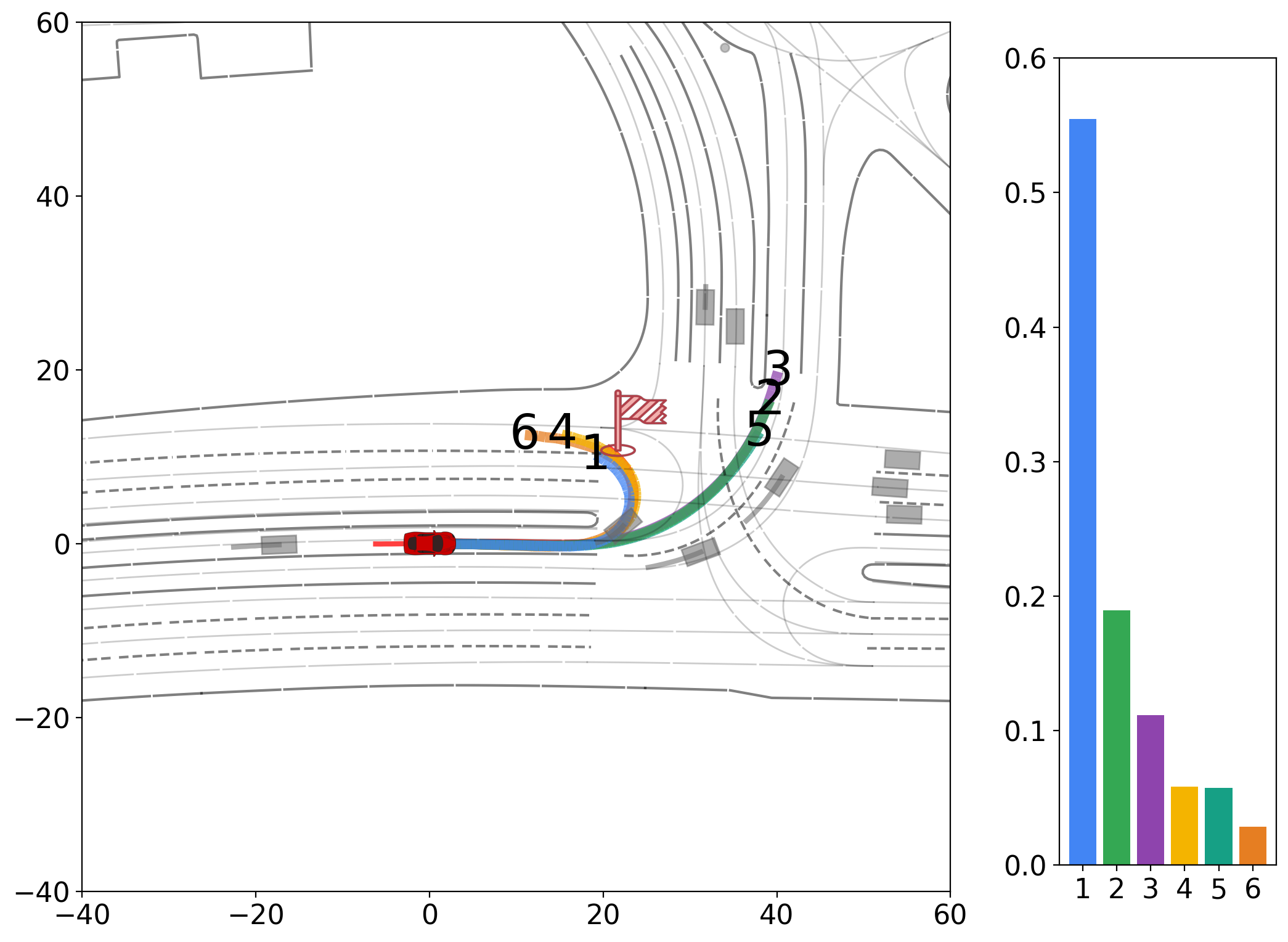}
    \includegraphics[width=0.48\textwidth]{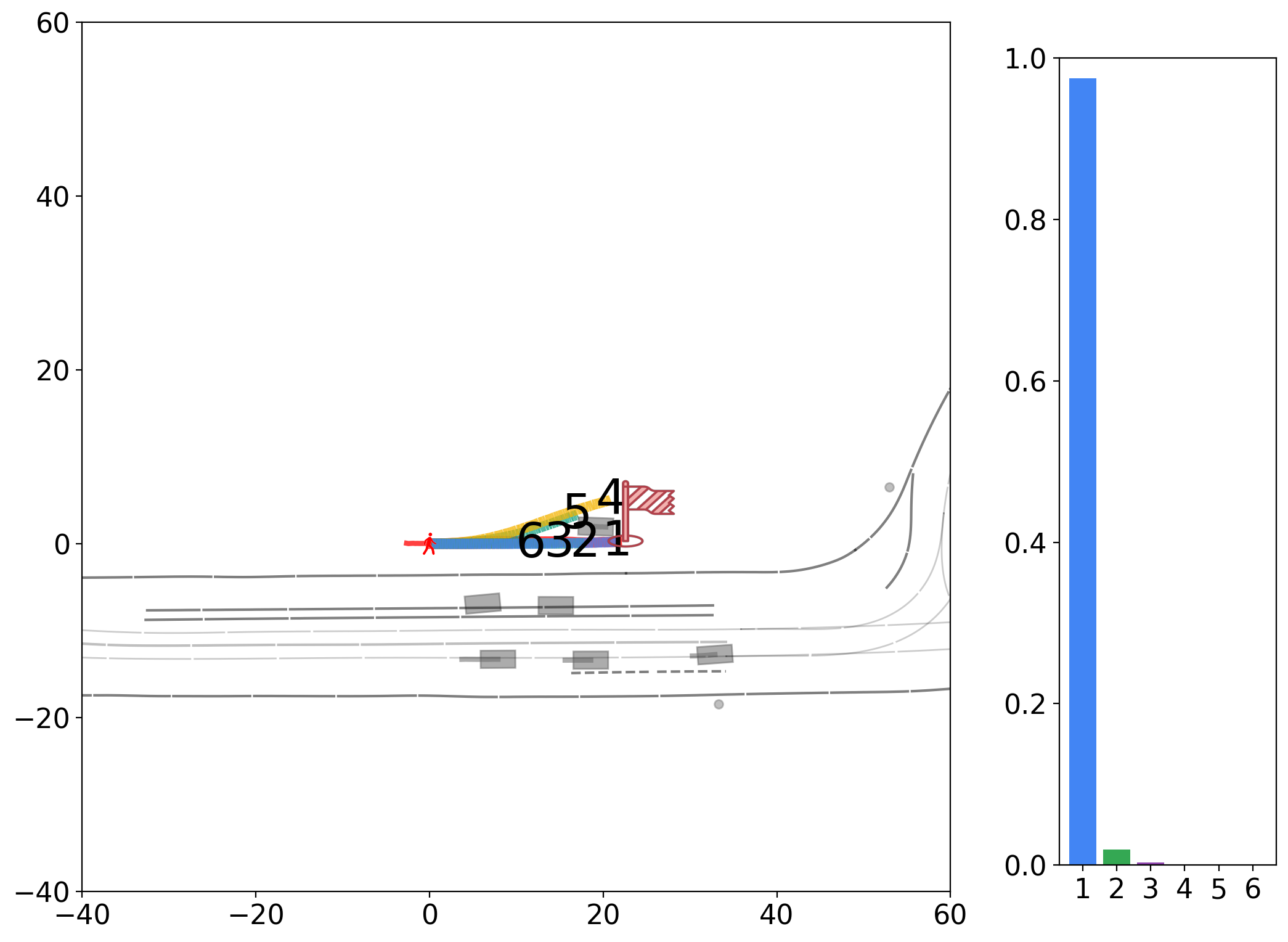}
    \includegraphics[width=0.48\textwidth]{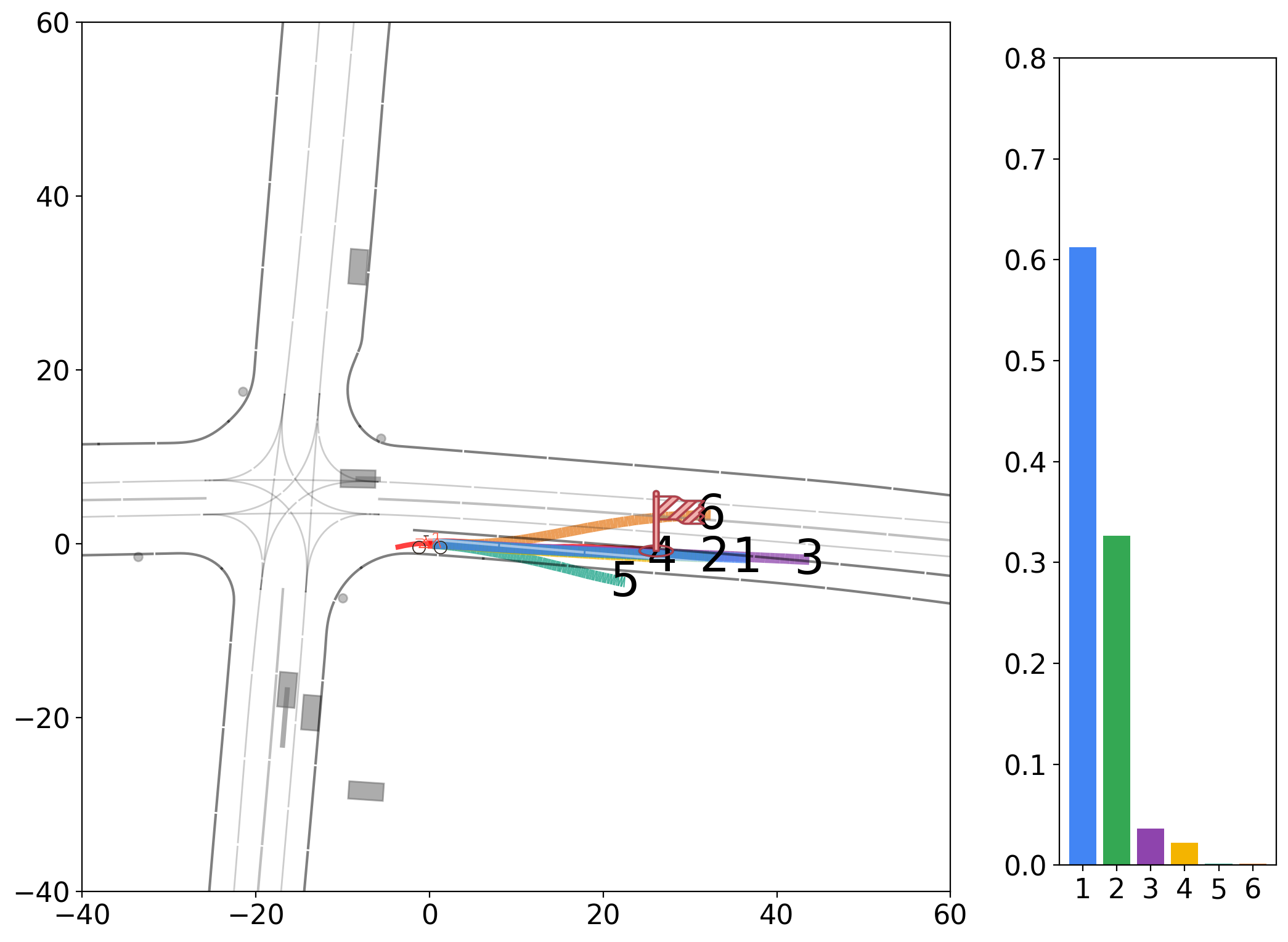}
    \includegraphics[width=0.48\textwidth]{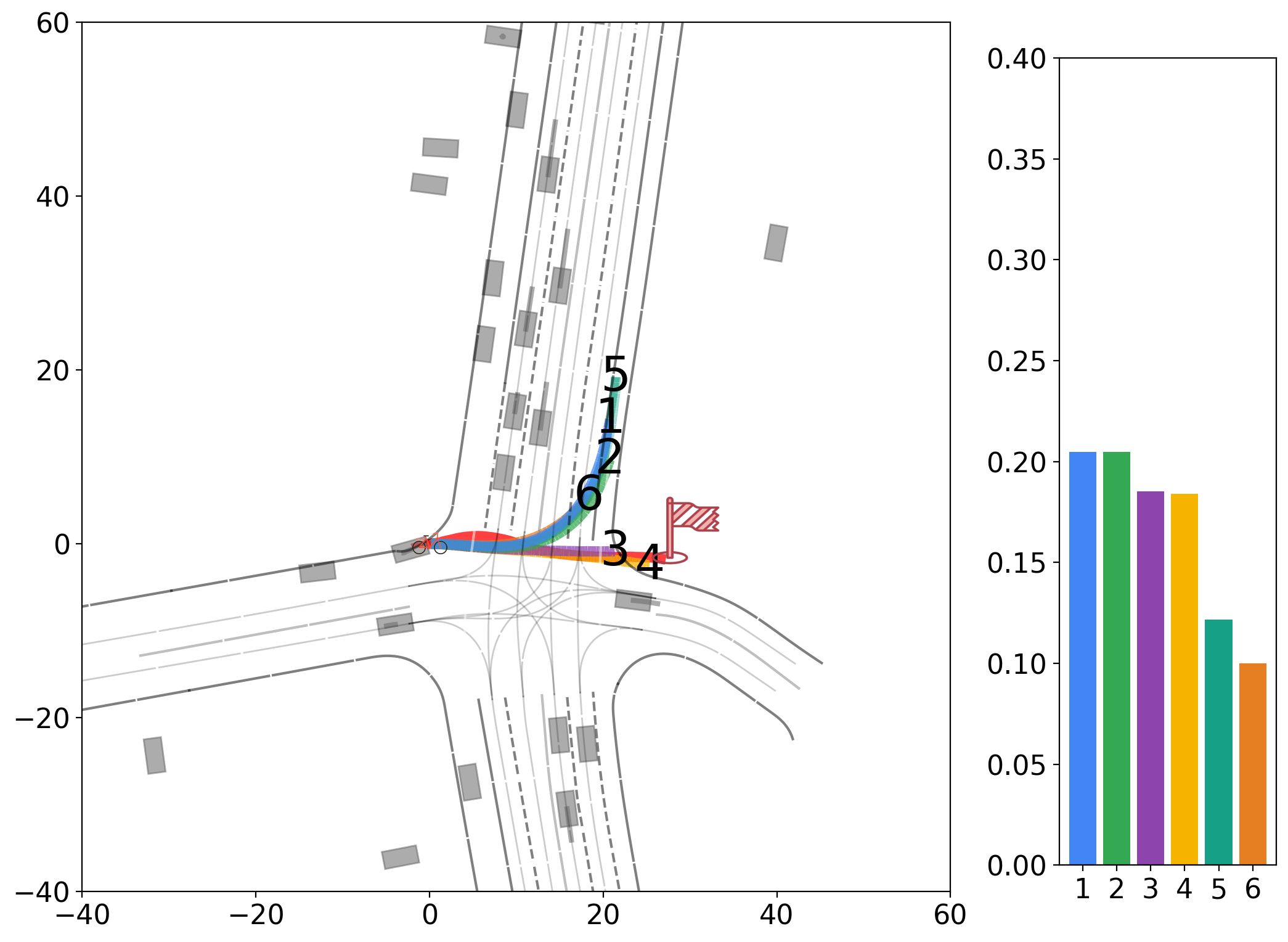}
    \caption{
    \textbf{Additional qualitative results.} 
    We visualize trajectory predictions (left) and their normalized confidence scores (right) in each plot. Six trajectory predictions are marked with different colors and numbered at their endpoints; the most confident prediction is shown in blue. The ground truth is displayed in red, with a flag indicating its endpoint. Surrounding road elements, such as lanes and sidewalks, are shown as lines and dots. Other surrounding traffic objects are represented as gray boxes.
    }
    \label{fig:traj_success_more}
\end{figure*}

\begin{figure*}[htbp]
    \centering
    \includegraphics[width=0.48\textwidth]{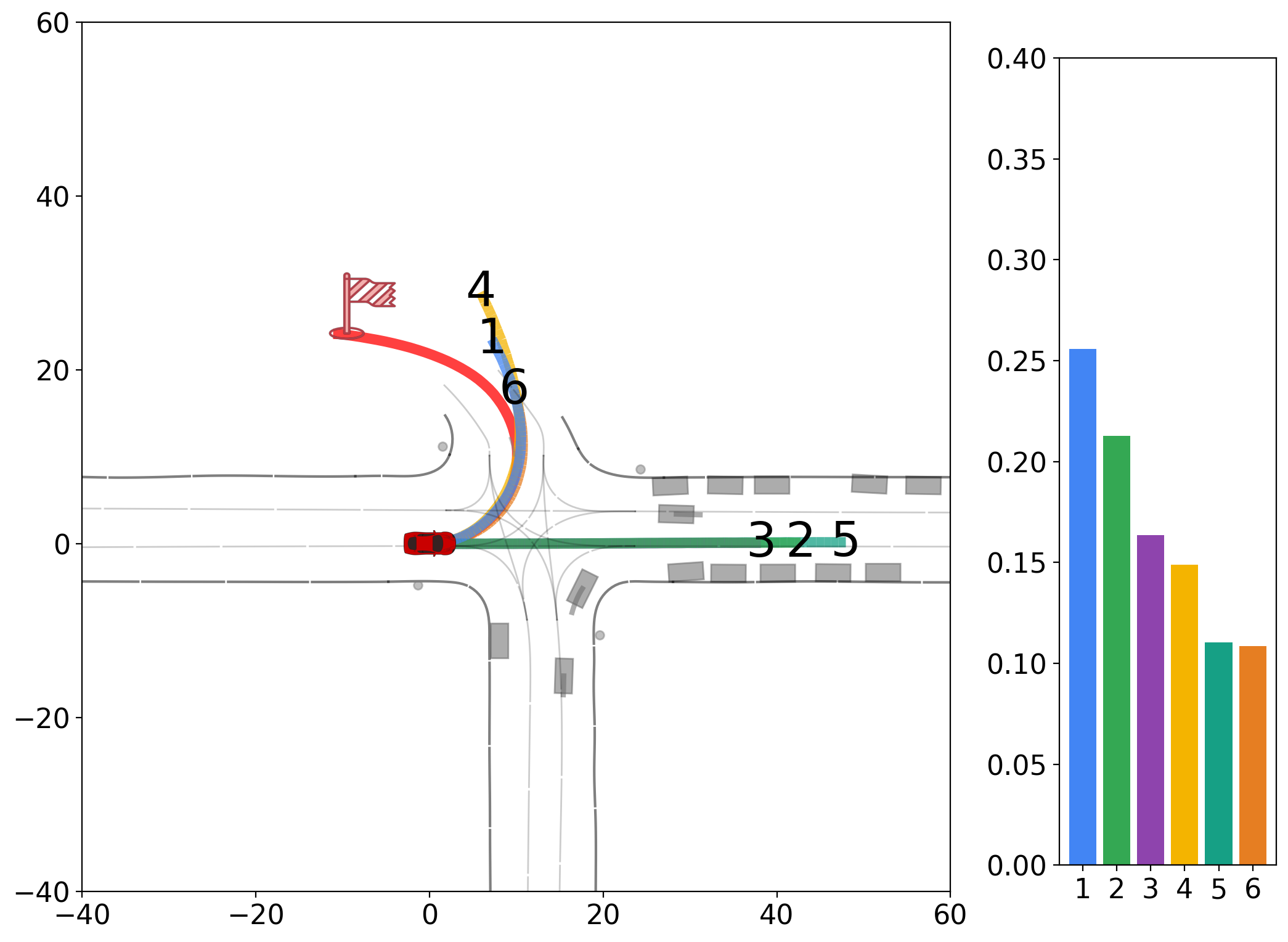} 
    \includegraphics[width=0.48\textwidth]{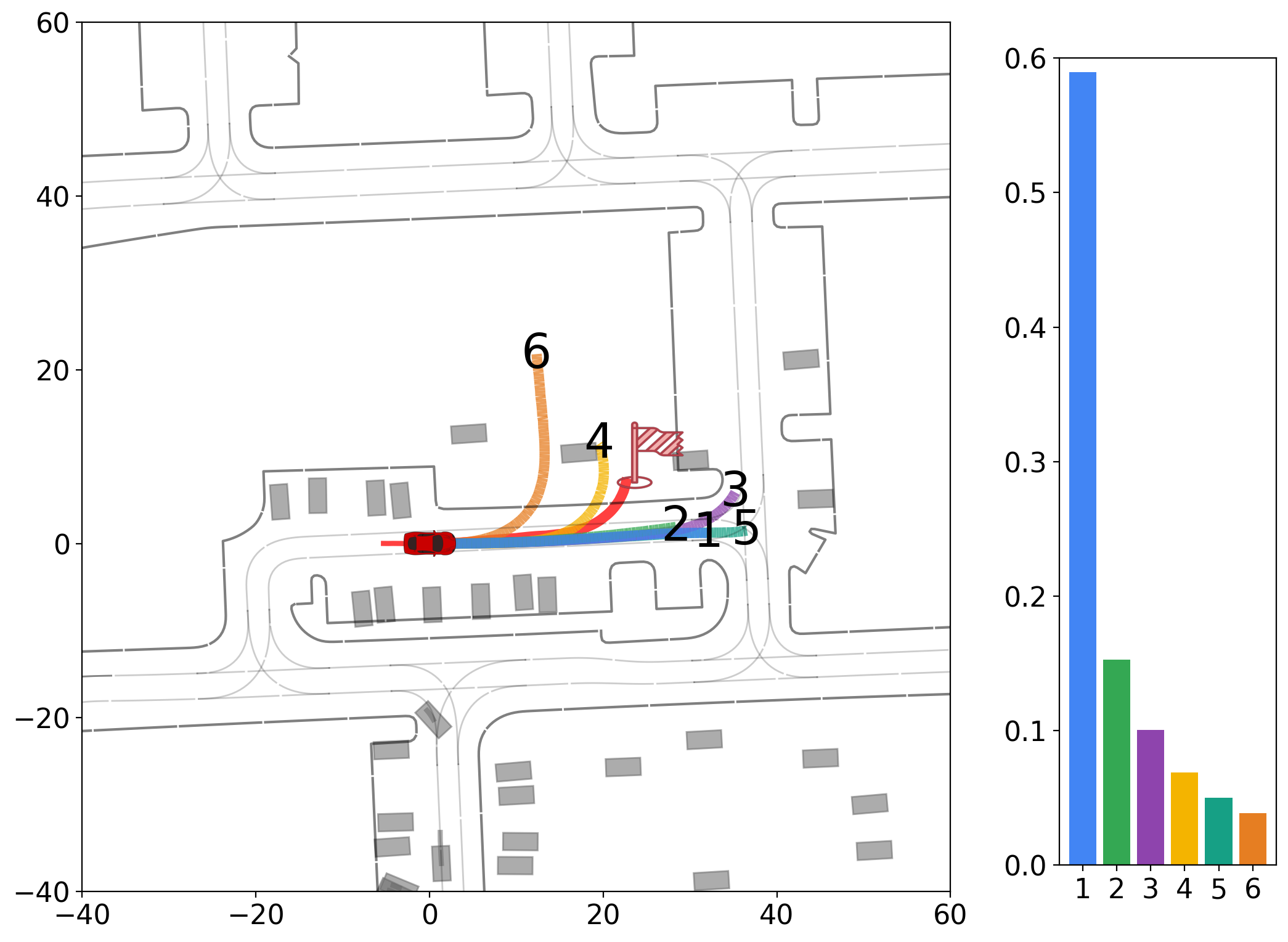}
    \includegraphics[width=0.48\textwidth]{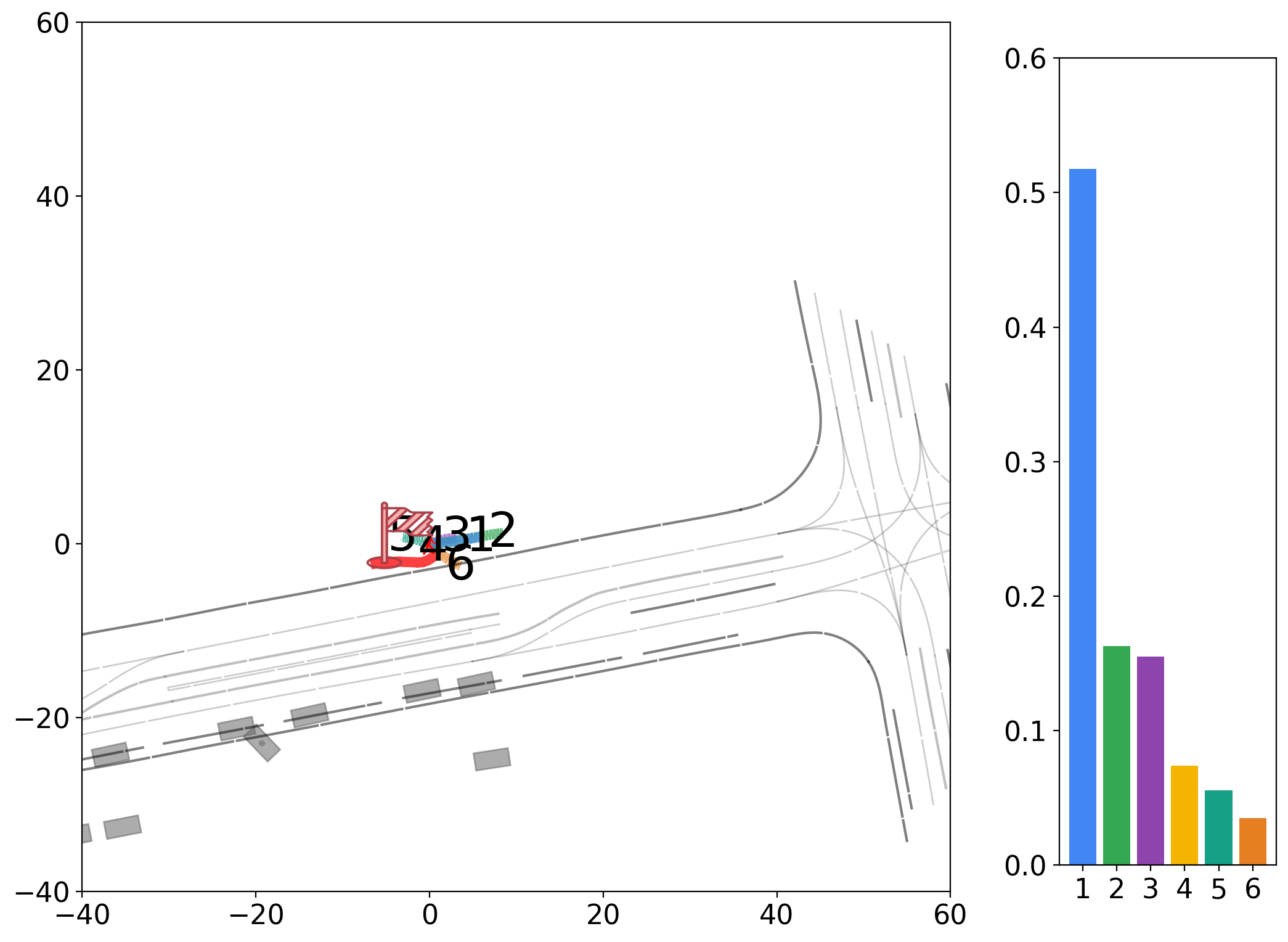}
    \includegraphics[width=0.48\textwidth]{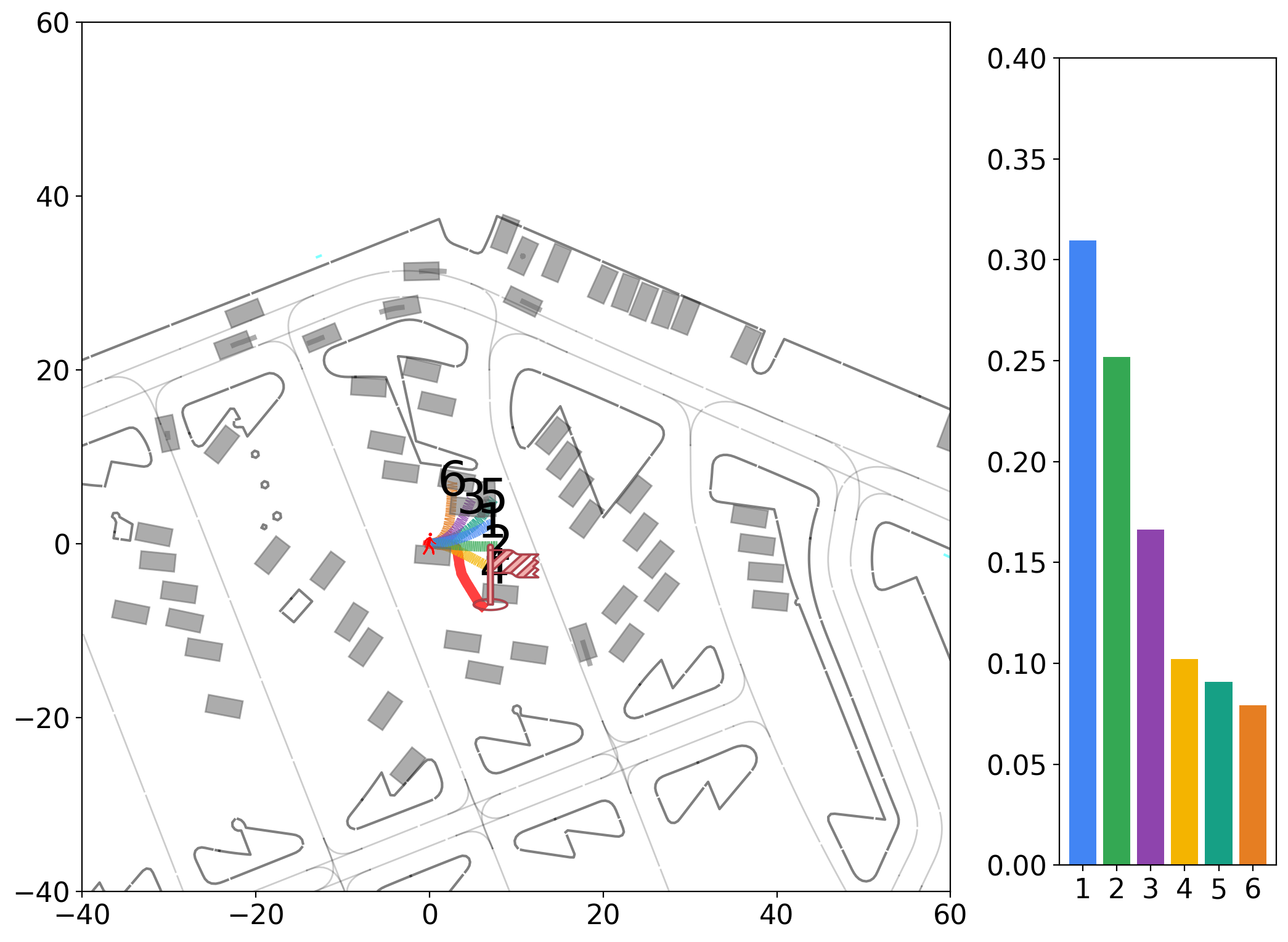}
    \caption{
    \textbf{Failure cases}. Given the volatility of real-world traffic, the models may not always predict accurate waypoints.
    }
    \label{fig:traj_failure}
\end{figure*}
\subsubsection{More Qualitative Results}
Additional qualitative results are presented in~\cref{fig:traj_success_more} and~\cref{fig:traj_failure}, providing further examples of both successful and failed cases on the validation datasets.
In~\cref{fig:traj_success_more}, we showcase a diverse set of agents—including vehicles, pedestrians, and cyclists—making reasonable predictions. In these cases, at least one of the predicted trajectories closely aligns with the ground truth, while the remaining predictions exhibit diverse and plausible motion patterns. This highlights the model’s ability to capture multi-modality and generate varied hypotheses about future movements.

Inaccurate predictions can be attributed to two main factors: (1) the predicted trajectories may deviate significantly from the ground truth in space, leading to prediction errors; and (2) the most accurate trajectories may not receive the highest confidence scores, resulting in calibration errors. The last row of~\cref{fig:traj_success_more} highlights two special cases where trajectories close to the ground truth are present, but are not ranked highest, indicating a calibration issue. Despite the overall strong performance of the proposed~\modelName model, it struggles in certain challenging scenarios, as shown in~\cref{fig:traj_failure}, where predictions spatially diverge from the ground truth.
These failure cases often involve complex interactions, rare or unseen motion patterns, or ambiguous intent, all of which can challenge and mislead the model. We attribute such limitations to a combination of data scarcity in these edge-case scenarios and the inherent uncertainty of real-world driving, which demands a nuanced understanding of contextual cues and agent intent.

\subsubsection{Societal Impact}
Our method supports motion prediction for agents in autonomous driving scenarios, contributing to safer navigation and improved decision-making. However, its reliability in edge cases—such as rare behaviors or complex interactions—remains an open question. For safety-critical deployment, these limitations must be addressed through rigorous testing, robust uncertainty estimation, and careful consideration of potential failure modes. Broader societal implications, including fairness across agent types and accountability in decision-making, should also be carefully considered.

\end{document}